\documentclass{article}

\usepackage{microtype}
\usepackage{graphicx}
\usepackage{subfigure}
\usepackage{booktabs} 
\usepackage{amsmath}
\usepackage{amssymb}
\usepackage{mathtools}
\usepackage[table]{xcolor}   
\usepackage{booktabs}        
\usepackage{multirow}        
\usepackage{pifont}
\usepackage{tabularx}
\usepackage{makecell}
\usepackage{array}

\newcommand{\ballotx}{\textcolor{red}{\ding{55}}}   

\newcolumntype{L}{>{\raggedright\arraybackslash}p{0.09\textwidth}}
\newcolumntype{M}{>{\centering\arraybackslash\scriptsize}X}
\definecolor{headergray}{RGB}{245,245,245}
\newcommand{\multfac}[1]{\textcolor{green!50!black}{(×#1)}}
\newcommand{\multfacblack}[1]{\textcolor{black!70}{(×#1)}}
\newcommand{\metriccell}[2]{\makecell{#1\\{\color{blue}(#2)}}}
\newcolumntype{C}[1]{>{\centering\arraybackslash}m{#1}}
\newcolumntype{V}{>{\raggedleft\arraybackslash}p{0.48\columnwidth}}

\usepackage{hyperref}

\usepackage[accepted]{mlsys2025}

\mlsystitlerunning{CDLM: Consistency Diffusion Language Models for Faster Sampling}

\begin{document}

\newcommand{\coleman}[1]{\textcolor{red}{[Coleman: #1]}}
\newcommand{\amir}[1]{\textcolor{red}{[Amir: #1]}}
\newcommand{\michael}[1]{\textcolor{red}{[Michael: #1]}}
\newcommand{\minseo}[1]{\textcolor{blue}{[Minseo: #1]}}
\newcommand{\cfx}[1]{\textcolor{darkgreen}{Chenfeng: #1]}}
\newcommand{\hs}[1]{\textcolor{purple}{[Harman: #1]}}

\twocolumn[
\mlsystitle{CDLM: Consistency Diffusion Language Models \\ for Faster Sampling}



\mlsyssetsymbol{equal}{*}

\begin{mlsysauthorlist}
\mlsysauthor{Minseo Kim}{snu}
\mlsysauthor{Chenfeng Xu}{ucb,together}
\mlsysauthor{Coleman Hooper}{ucb}
\mlsysauthor{Harman Singh}{ucb}
\mlsysauthor{Ben Athiwaratkun}{together}
\mlsysauthor{Ce Zhang}{together}
\mlsysauthor{Kurt Keutzer}{ucb}
\mlsysauthor{Amir Gholami}{ucb}
\end{mlsysauthorlist}

\mlsysaffiliation{snu}{Seoul National University, Seoul, South Korea}
\mlsysaffiliation{ucb}{University of California, Berkeley, CA, USA}
\mlsysaffiliation{together}{Together AI, San Francisco, CA, USA}

\mlsyscorrespondingauthor{Amir Gholami}{amirgh@berkeley.edu}

\mlsyskeywords{Machine Learning, Efficiency, AI Systems}

\vskip 0.3in

\begin{abstract}
Diffusion Language Models (DLMs) offer a promising parallel generation paradigm but suffer from slow inference due to numerous refinement steps and the inability to use standard KV caching.
We introduce CDLM (Consistency Diffusion Language Models), a training-based acceleration method that simultaneously tackles both bottlenecks.
CDLM integrates consistency modeling to drastically reduce the number of required sampling steps by enabling multi-token finalization.
Furthermore, we enforce a block-wise causal attention mask during fine-tuning, making the model fully compatible with KV caching.
Experiments show CDLM achieves $3.6\times$--$14.5\times$ lower latency while maintaining competitive accuracy on math and coding tasks.
The full training and evaluation code is available at
\href{https://github.com/SqueezeAILab/CDLM}{\texttt{https://github.com/SqueezeAILab/CDLM}}.
\end{abstract}
]



\printAffiliationsAndNotice{}  

\section{Introduction}
\label{introduction}

Large Language Models (LLMs) have triggered a paradigm shift in AI, excelling at tasks such as instruction following, multi-turn dialogue, and code generation~\cite{openai2023gpt4}.
Most current LLMs are autoregressive, decoder-only Transformers trained to predict the next token under a causal attention mask~\cite{grattafiori2024llama, zhang2022opt}.
While training is parallelizable via teacher forcing, inference inherits a rigid sequential dependency, becoming a key efficiency bottleneck.
Moreover, such models cannot exploit bidirectional context, which limits performance on tasks such as text infilling, editing, and global planning~\cite{hu2024bai, song2025seeddiffusion}.

As a promising alternative, diffusion language models (DLMs) have drawn attention for their potential in text generation.
Inspired by the success of diffusion models in image, audio, and video synthesis~\cite{ho2020ddpm}, DLMs iteratively denoise random or masked token sequences into coherent text~\cite{gong2023diffuseq, li2022diffusionlm}.
This process updates all token positions \textit{in parallel} per step, removing the token-level sequential dependency inherent to autoregressive decoding.
Closed-source models, primarily in code generation, report up to $10\times$ higher throughput than autoregressive models while preserving quality~\cite{deepmind2025geminidiffusion,khanna2025mercury,song2025seeddiffusion}.
However, open-source DLMs remain significantly slower~\cite{nie2025large, ye2025dream7b}, largely because (i) bidirectional attention prohibits standard KV caching and (ii) they demand many refinement steps, often proportional to sequence length, to reach high fidelity~\cite{kim2025beyond}.

Many recent works aim to accelerate DLM inference.
One primary strategy leverages block-wise decoding with KV caching, either via approximate caching with periodic updates~\cite{hu2025freecache, liu2025dllmcache, ma2025dkvcache, wu2025fastdllm} or by fine-tuning the model into a block-wise causal architecture so that caching becomes natural~\cite{wang2025d2f, wu2025fastdllmv2}.
The other major line of work focuses on parallel sampling algorithms, which effectively decode multiple tokens per denoising step and reduce the total number of iterations~\cite{benhamu2025ebsampler, wu2025fastdllm}.

In diffusion models for vision domains, many works have successfully reduced denoising to few-step or even one-step generation~\cite{yin2024dmd, yin2025causvid}.
A key technique behind this progress is \emph{consistency modeling}, which trains models to produce consistent predictions across adjacent denoising states along the probability flow trajectory~\cite{song2023consistency, song2023improved}.
Although originally developed for continuous data, this paradigm has begun to influence discrete settings, including parallel decoding for autoregressive language models~\cite{kou2024cllms} and masked diffusion frameworks~\cite{xu2025unicms}, highlighting its promise for language diffusion models.

In this work, we present \textbf{CDLM}, a training-based acceleration method that integrates \textbf{C}onsistency modeling~\cite{song2023consistency, song2023improved} into \textbf{D}iffusion \textbf{L}anguage \textbf{M}odels.
CDLM trains a block-causal student via distillation from a bidirectional teacher to finalize multiple tokens per refinement step, and applies confidence-thresholded parallel decoding within each block at inference time.
This design directly addresses two core bottlenecks of current DLMs:
(i) excessive refinement steps (mitigated by distillation and consistency) and
(ii) cache inefficiency (removed by block-wise causality).
In particular, we make the following contributions:

\vspace{-1mm}
\begin{itemize}
  \item \textbf{Consistency-guided distillation.}
  We bring consistency modeling to DLMs by defining a token-level decoding trajectory and coupling it with distillation from a fully bidirectional teacher.
  This joint objective provides explicit supervision for multi-token finalization while aligning predictions between a less-informed state and a block-completion state within each block (Sections~\ref{preliminary},~\ref{subsec:training}).
  \vspace{-0.5mm}
  \item \textbf{Cache-friendly block causality.}
  By fine-tuning with a block-wise causal attention mask (Figure~\ref{fig:attn-masks}), we make bidirectional DLMs compatible with block-wise KV caching and early stopping at block boundaries ( Sections~\ref{subsec:trajectory-collection},~\ref{subsec:inference}).
  We also provide a detailed system-level analysis of block-causal DLMs to characterize their efficiency and scalability (Section~\ref{subsubsec:system-analysis}).
  \vspace{-0.5mm}
  \item \textbf{End-to-end speedups.}
  With 8 hours of training for Dream and 16 hours for LLaDA, CDLM reduces refinement steps by $3.4{\times}$--$7.9{\times}$ and latency by $3.6{\times}$--$14.5{\times}$ while maintaining accuracy across math and coding benchmarks.  
  It also surpasses equal-size autoregressive LLMs in tokens-per-second (Section~\ref{subsec:main-results}).
\end{itemize}
\vspace{-1mm}
\begin{figure*}[t]
  \centering
  \includegraphics[width=\textwidth]{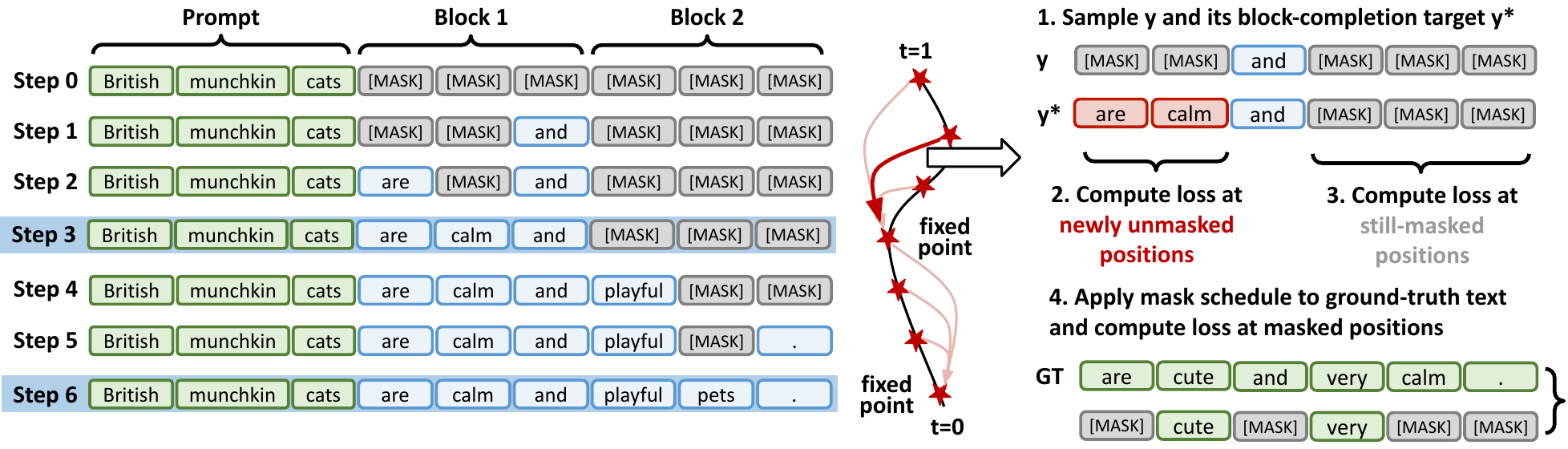}
  \vspace{-7mm}
  \caption{
  \textbf{Left:} Block-wise decoding trajectory of the teacher (steps $0\!\to\!N$; diffusion time $t:1\!\to\!0$).
\textbf{Right:} The student's three-objective loss at an intermediate state $y$:
(i) distillation from teacher logits on newly unmasked positions,
(ii) consistency between $y$ and its block-completion $y^{\star}$, and
(iii) masked-denoising (DLM) loss on randomly masked ground-truth text.
  }
  \label{fig:trajectory}
\end{figure*}

\section{Background}
\label{background}

\subsection{Diffusion Language Models}

Diffusion models~\cite{ho2020ddpm} have been successfully applied across modalities such as text-conditioned image generation~\cite{ramesh2022hierarchical}, audio synthesis~\cite{chen2021wavegrad}, and video generation~\cite{esser2023sceneguided}. 
Early Diffusion Language Models (DLMs) operated in the continuous embedding space~\cite{gao2024embedding, gong2023diffuseq}, but more recent models adopt masked diffusion models (MDMs) that work directly in the discrete token space~\cite{gong2025scalingdiffusion, nie2025scalingmdm}. These models begin from a fully masked sequence and iteratively unmask tokens according to a scheduling policy.
Architecturally, DLMs commonly employ Transformer backbones with \textit{bidirectional} (non-causal) attention, permitting full-context interactions~\cite{nie2025large}. Despite their parallel update capability, DLMs are known to be slow for two main reasons.
First, full bidirectional attention at every denoising step makes DLMs incompatible with standard KV caching.
Second, high-quality generation demands a large number of denoising steps comparable to the sequence length, making inference inefficient~\cite{kim2025beyond}.
In this work, we propose a fine-tuning-based solution that addresses both limitations.

\subsection{Efficient Inference in Diffusion Language Models}

Inference in DLMs is an active area of research with significant room for improvement.
Among training-free acceleration methods, one line of work exploits approximate caching via block-wise decoding, only recomputing the active block while reusing stale key–value states for inactive blocks~\cite{ma2025dkvcache, wu2025fastdllm}.
Others speed up parallel sampling by applying confidence thresholds~\cite{wu2025fastdllm} or by leveraging lightweight autoregressive models as guidance~\cite{hu2025freecache}, while recent analysis suggests that effective parallelism is difficult to adapt reliably at inference time~\cite{kang2026parallelbench}.
In contrast, training-based approaches focus on improving the intrinsic efficiency of DLMs through architectural or objective-level modifications.
Some recent works fine-tune pretrained AR models or DLMs to introduce block-wise causality, enabling standard KV caching at the block level while retaining diffusion-style updates within each block~\cite{wang2025d2f, wu2025fastdllmv2}.
Others design objectives that promote faster convergence by encouraging high-confidence token prediction~\cite{chen2025dparallel}.  
Our method is training-based: we jointly induce block-wise causality and reduce refinement steps through consistency-guided fine-tuning.

\subsection{Knowledge Distillation in Language Models}

Knowledge distillation (KD)~\cite{hinton2015distilling} transfers knowledge from a large, powerful teacher to a smaller, more efficient student.  
Early work on LLM distillation primarily relied on \textit{black-box} KD, where students imitated textual outputs from closed-source APIs~\cite{fu2023specializing}.  
With the rise of open-source LLMs, \textit{white-box} distillation has become prevalent, enabling richer supervision from the teacher’s internal representations, such as logits or embeddings~\cite{agarwal2024onpolicy, gu2023knowledge}.
Beyond language, distillation has also been applied to diffusion-based image and video generation to reduce sampling steps while preserving quality~\cite{yin2024dmd, yin2025causvid}.
Our method can similarly be viewed as a form of \textit{self-distillation}, where knowledge is transferred between models of identical size and type but with different architectural designs (a fully bidirectional teacher and a block-wise causal student).
Like \citet{deschenaux2025sdtt} and \citet{xu2025dinfer}, we reduce decoding steps by distilling multi-step diffusion trajectories into \textit{jumps} between non-consecutive denoising states.
Our novelty lies in instantiating these jumps within a block-causal student and training it with \emph{consistency}-oriented objectives.

\subsection{Consistency Models and Language Models}

Consistency models~\cite{song2023consistency} were introduced to overcome the slow, iterative sampling process of diffusion models.  
They learn a \emph{consistency} property: any intermediate state along a probability flow ODE trajectory can be mapped directly back to its origin (the clean data)~\cite{song2023improved}.
This is achieved by minimizing discrepancies between predictions at adjacent denoising states.
While originally developed for continuous domains such as vision, recent work has explored adapting this idea to discrete language modeling~\cite{kou2024cllms, xu2025unicms}.
Rather than explicitly modeling the ODE solver or its continuous trajectory, these methods reinterpret the notion of mapping intermediate states to final solutions within discrete token trajectories~\cite{santilli2023accelerating}.
Our approach adopts this paradigm in DLMs, where trajectories arise naturally from iterative denoising.
Conceptually, our objective can be viewed as an empirical generalization of the original consistency objective defined over continuous ODE trajectories, adapted to discrete, token-level diffusion processes.

\section{Preliminary: Inference in Diffusion Language Models}
\label{preliminary}

We formalize DLM inference by defining the iterative refinement process, the sampling strategies, and the decoding trajectory that underpins our training objectives.

\paragraph{Iterative Refinement Process.}

DLM generation is an iterative refinement over $N$ discrete sampling steps.
It gradually transforms a fully masked sequence $\mathbf{x}_1$ at time $t=1$ \,(step $0$)\, into a clean sequence $\mathbf{x}_0$ at $t=0$ \,(step $N$).
Here, $N$ serves as a hyperparameter controlling the trade-off between generation speed and sample quality.
We denote the sequence at a discrete step $k$ (where $k = 0,1,\dots,N$) as $\mathbf{x}_{t_k}$ with
$t_k = 1 - \frac{k}{N}$ (so $t_0{=}1$, $t_N{=}0$).
At each step, approximately $L_g / N$ tokens are finalized on average, where $L_g$ is the target generation length.

This refinement process is modeled probabilistically. 
Specifically, the transition from a state $\mathbf{x}_t$ to a subsequent state $\mathbf{x}_s$ $0 \le s < t \le 1$ is governed by the model $p_\theta$'s prediction of the clean sequence $\mathbf{x}_0$ given the current noisy sequence $\mathbf{x}_t$ and a conditioning prompt $c$, i.e.,
\begin{equation}
    p_\theta(\mathbf{x}_0 \mid \mathbf{x}_t, c).
\end{equation}
Following prior formulations~\cite{austin2021structured}, the reverse-time transition
$q_{s|t}(\mathbf{x}_s \mid \mathbf{x}_t)$ is defined for $0 \le s < t \le 1$ and factorizes across tokens.
For each token $i$, the transition probability is defined as:
\begin{equation}
q_{s|t}(x_s^i \mid x_t^i) =
\begin{cases}
1, & \\
\quad\quad\text{if } x_t^i \neq \texttt{[MASK]},\; x_s^i = x_t^i, \\[6pt]
\frac{s}{t}, & \\
\quad\quad\text{if } x_t^i = \texttt{[MASK]},\; x_s^i = \texttt{[MASK]}, \\[6pt]
\frac{t-s}{t}\, q_{0|t}(x_s^i \mid \mathbf{x}_t, c), & \\
\quad\quad\text{if } x_t^i = \texttt{[MASK]},\; x_s^i \neq \texttt{[MASK]}.
\end{cases}
\end{equation}

where \(q_{0|t}\) is the predictive distribution induced by \(p_\theta(\mathbf{x}_0 \mid \mathbf{x}_t, c)\).
This formulation reflects three token-level transitions: preserving an unmasked token, keeping a token masked, or unmasking it to a new token.

\paragraph{Sampling Strategies.}
To instantiate the probabilistic distribution $q_{s|t}$ into an actual generation procedure, practical deterministic strategies are employed. 
A common choice is \textit{low-confidence remasking}, where instead of stochastic sampling, the model greedily selects the tokens to unmask~\cite{nie2025large}.  
Based on the confidence scores from $p_\theta(\mathbf{x}_0 \mid \mathbf{x}_t, c)$, the top-$m$ masked tokens with the highest probabilities are revealed, while all others remain masked. 
In practice, this process is often constrained to operate within \textit{blocks} of tokens, called \textit{block-wise decoding}, limiting unmasking to localized regions of the sequence.  
This prevents unnatural early generation~\cite{nie2025large} and, more importantly, enables KV caching for finalized blocks.
Parallel denoising implicitly assumes \textit{independence} among simultaneously finalized tokens, an idealized assumption that can degrade generation quality as more tokens are unmasked at once~\cite{liu2025tidar}.

\paragraph{Decoding Trajectory.}
Given this iterative and probabilistic refinement process, we define the \textit{decoding trajectory} $\mathcal{T}$ as the sequence of states that the model traverses during inference:
\begin{equation}
\mathcal{T}_{\textbf{x}}
= \big(\mathbf{x}_{t_0},\, \mathbf{x}_{t_1},\, \ldots,\, \mathbf{x}_{t_N}\big),
\quad t_k = 1 - \frac{k}{N}.
\end{equation}

Each \(\mathbf{x}_{t_k}\) is a partially refined sequence;
the trajectory \(\mathcal{T}_{\mathbf{x}}\) records its evolution across steps. Figure~\ref{fig:trajectory} (left) visualizes an example.
This trajectory serves as the foundation for our consistency-based training below.

\section{Methodology}
\label{methodology}

We present \textbf{Consistency Diffusion Language Models (CDLM)}, a training scheme that accelerates DLM inference without sacrificing quality.
We describe trajectory and hidden-state collection, our three-objective training, and the inference strategy.

\begin{figure}[t]
  \centering
  \includegraphics[width=\linewidth]{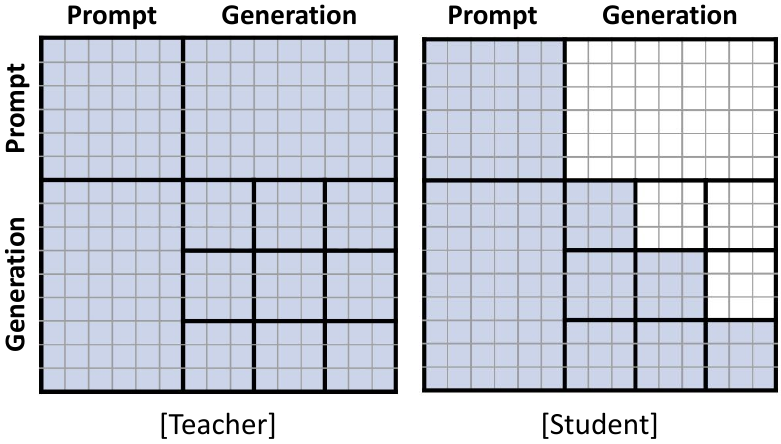}
  \vspace{-7mm}
  \caption{
  \textbf{Left:} Teacher DLM with full bidirectional attention, attending to the entire context.
  \textbf{Right:} Student DLM with a block-wise causal mask, attending to the prompt, previously completed blocks, and the current decoding block.
  }
  \label{fig:attn-masks}
\end{figure}

\begin{algorithm}[t]
\caption{Trajectory Collection for Training CDLM}
\label{alg:traj-collection}
\begin{algorithmic}[1]
\STATE \textbf{Input:} dataset $\mathcal{O}$ of pairs $(x,\hat{\mathbf{y}})$, generation length $L_g$, block size $B$, total steps $N{=}L_g$, teacher DLM $p_\theta$, temperature set $\mathcal{S}_\tau$, target count $n$
\STATE \textbf{Output:} dataset $\mathcal{D}$ of quadruples $(x, \hat{\mathbf{y}}, \mathcal{T}_x, \mathbf{H}_x)$
\STATE Initialize $\mathcal{D}\leftarrow\emptyset$
\REPEAT
  \STATE Sample $(x,\hat{\mathbf{y}})\sim\mathcal{O}$; build input ids $\mathbf{x}$ with the answer span masked
  \FOR{each temperature $\tau\in\mathcal{S}_\tau$}
    \STATE $\mathcal{T}\leftarrow[\,]$;\; $\mathbf{H}\leftarrow\mathbf{0}\in\mathbb{R}^{L_g\times d}$
    \FOR{blocks $b=1,\dots,\lceil L_g/B\rceil$}
      \FOR{steps $s=1,\dots,B$}
        \STATE Run $p_\theta(\mathbf{x})$; obtain logits $\ell$ and last hidden $\mathbf{h}$
        \STATE Sample $\mathbf{y}$ and compute confidence $\mathbf{c}$
        \STATE Finalize top-1 token $y_i$ in the current block; write $\mathbf{h}_i\!\to\!\mathbf{H}$; append $\mathbf{x}\!\to\!\mathcal{T}$
      \ENDFOR
    \ENDFOR
    \STATE Append $(x,\hat{\mathbf{y}},\mathcal{T},\mathbf{H})$ to $\mathcal{D}$
  \ENDFOR
\UNTIL{$|processed\_pairs|\ge n$}
\end{algorithmic}
\end{algorithm}

\subsection{Trajectory Collection for CDLM}
\label{subsec:trajectory-collection}

Let $p_{\theta}$ denote the teacher DLM and $q_{\phi}(\cdot \mid x)$ the student DLM initialized from the teacher’s weights.
The teacher uses a \textit{bidirectional} attention mask as in standard DLMs, whereas the student is trained with a \textit{block-wise causal} mask, as illustrated in Figure~\ref{fig:attn-masks}.
To enable consistency training, we collect trajectories offline by running the teacher on domain-specific prompts.
For each prompt $x$, we denote the token-level trajectory as $\mathcal{T}_x$ and the corresponding hidden-state buffer by $\mathbf{H}_x \in \mathbb{R}^{L_g\times d}$; the resulting dataset is $\mathcal{D}=\{(x,\hat{\mathbf{y}},\mathcal{T}_x,\mathbf{H}_x)\}$ with ground-truth text $\hat{\mathbf{y}}$ included.
The overall procedure is summarized in Algorithm~\ref{alg:traj-collection}.

Prior work reports that block-wise decoding is effective for instruction-tuned models and attains high quality when the number of steps matches the generation length~\cite{nie2025large}.
Accordingly, we adopt block-wise decoding with the number of sampling steps equal to the generation length ($N{=}L_g$) and finalize exactly one token per step within the current block (the highest-confidence token $y_i$).
This configuration places the teacher at its most performant operating point and yields higher-quality trajectories.
For white-box distillation, we store not only $\mathcal{T}_x$ but also the states required to reconstruct logits.
Concretely, we record the last hidden states at the moments when tokens are finalized. We also augment each prompt with multiple trajectories generated at different temperatures.
Additional details on dataset construction are provided in Appendix~\ref{appendix:dataset}.

\begin{algorithm}[t]
\caption{Training Algorithm for CDLM}
\label{alg:cdlm-train}
\begin{algorithmic}[1]
\STATE \textbf{Input:} dataset $\mathcal{D}=\{(x,\hat{\mathbf{y}},\mathcal{T}_x,\mathbf{H}_x)\}$, block size $B$, generation length $L_g$, weights $w_{\text{cons}}, w_{\text{distill}}$, $w_{\text{dlm}}$, student DLM $q_\phi(\cdot\mid x)$ \emph{with block-wise causal attention}, total steps $N{=}L_g$
\STATE \textbf{Output:} updated student parameters $\phi$
\REPEAT
  \STATE Sample $(x,\hat{\mathbf{y}},\mathcal{T}_x,\mathbf{H}_x)\sim\mathcal{D}$
  \STATE Sample $t_{\text{start}}$; set $t_{\text{end}}\leftarrow \min\!\left(N,\, \left\lceil \tfrac{t_{\text{start}}}{B} \right\rceil B\right)$
  \STATE Retrieve states at $t_{\text{start}}$ and $t_{\text{end}}$ from $\mathcal{T}_x$
  \STATE Compute $\mathcal{L}_{\text{Distillation}}$ (Eq.~(4)) using $\mathbf{H}_x$
  \STATE Compute $\mathcal{L}_{\text{Consistency}}$ (Eq.~(5))
  \STATE Compute $\mathcal{L}_{\text{DLM}}$ (Eq.~(6))
   using ground-truth text $\hat{\mathbf{y}}$
  \STATE $\mathcal{L} \leftarrow 
  w_{\text{distill}}\mathcal{L}_{\text{Distillation}} +
  w_{\text{cons}}\mathcal{L}_{\text{Consistency}} + 
  w_{\text{dlm}}\mathcal{L}_{\text{DLM}}$
  \STATE Update $\phi$ with $\mathcal{L}$
\UNTIL{max epochs}
\end{algorithmic}
\end{algorithm}

\subsection{CDLM Training}
\label{subsec:training}

We jointly minimize three objectives to train CDLMs:
(i) a \textbf{Distillation loss} that transfers the teacher’s multi-token finalization signal to the block-wise causal student;
(ii) a \textbf{Consistency loss} that enforces within-block temporal consistency between two states; and
(iii) a \textbf{DLM loss} that preserves the model’s masked-token prediction capability.
The detailed training procedure is depicted in Algorithm~\ref{alg:cdlm-train}.

\paragraph{Notation}

Given the dataset $\mathcal{D}=\{(x,\hat{\textbf{y}}, \mathcal{T}_x,\mathbf{H}_x)\}$, we first sample a prompt $x$ and its trajectory $\mathcal{T}_x$, then sample a state $y\in\mathcal{T}_x$ and its block-completion state $y^\star$.
Here, $y^\star$ denotes the state obtained by fully unmasking the active block of $y$ (thus $y$ and $y^\star$ are at most $B$ steps apart).
The right panel of Figure~\ref{fig:trajectory} visualizes this setup.
We denote $\mathcal{U}_y=\{i \mid y_i=\texttt{[MASK]},\, y_i^\star\neq\texttt{[MASK]}\}$ as the newly unmasked token indices between $y$ and $y^\star$, and $\mathcal{S}_y=\{i \mid y_i=\texttt{[MASK]},\, y_i^\star=\texttt{[MASK]}\}$ as the still-masked token indices at $y^\star$.

\paragraph{Distillation Loss}

We use the student’s predictions at state $y$ together with the teacher’s hidden-state buffer $\mathbf{H}_x$.
For each $i \in \mathcal{U}_y$, we reconstruct the teacher’s distribution by applying
$\ell^{(T)}_i = \texttt{lm\_head}(\mathbf{h}_{x,i})$ and
$p^{(T)}_i = \operatorname{softmax}(\ell^{(T)}_i)$
to the stored hidden state $\mathbf{h}_{x,i}$.
With these teacher distributions, we define the distillation objective as
\begin{equation}
\label{eq:kd}
\begin{aligned}
\mathcal{L}_{\text{Distillation}}
&= \mathbb{E}_{(x,\mathcal{T}_x,\mathbf{H}_x)\sim\mathcal{D}}\,\mathbb{E}_{y\sim\mathcal{T}_x}\Bigg[ \\[0.25em]
&\qquad \frac{1}{|\mathcal{U}_y|}
\sum_{i\in\mathcal{U}_y}
D_{\mathrm{KL}}\!\Big(
p^{(T)}_i \,\big\|\, q_\phi(\,\cdot \mid y, x)_i
\Big)
\Bigg],
\end{aligned}
\end{equation}
which applies forward KL divergence on the positions newly unmasked between $y$ and $y^\star$.
This serves as the main anchor that transfers the bidirectional teacher’s guidance to the block-wise student, effectively providing \textit{supervision for multi-token finalization} within each block.

\paragraph{Consistency Loss}

We compare the student’s predictive distributions at state $y$ and block-completion state $y^\star$, enforcing agreement on still-masked positions. Concretely, we minimize the following objective:
\begin{equation}
\label{eq:consistency}
\begin{aligned}
\mathcal{L}_{\text{Consistency}}
&= \mathbb{E}_{(x,\mathcal{T}_x)\sim\mathcal{D}}\,\mathbb{E}_{y\sim\mathcal{T}_x}\Bigg[\\
&\hspace{-1.5em}
\frac{1}{|\mathcal{S}_y|}
\sum_{i\in\mathcal{S}_y}
D_{\mathrm{KL}}\!\Big(
q_{\phi^-}(\,\cdot \mid y^\star, x)_i
\;\big\|\;
q_{\phi}(\,\cdot \mid y, x)_i
\Big)
\Bigg].
\end{aligned}
\end{equation}
Here, $q_{\phi^-}$ denotes a stop-gradient target (detached from backpropagation) for stable training~\cite{song2023consistency}, and $D_{\mathrm{KL}}$ is the forward KL divergence between two distributions.
Intuitively, Eq.~\eqref{eq:consistency} aligns the student at a \emph{less-informed} state with itself at a \emph{more-informed} state, encouraging stable multi-step jumps along the decoding trajectory.
Since DLMs are trained to predict only \texttt{[MASK]} tokens, restricting the loss to still-masked positions ($\mathcal{S}_y$) avoids ill-defined supervision.

\paragraph{DLM loss}

We include an auxiliary loss, identical to the masked denoising objective used in standard DLM pre-training~\cite{nie2025large,ye2025dream7b}, to preserve the model’s mask prediction capability.
For a prompt $x$ and its ground-truth response $\hat{\mathbf{y}}$, we randomly sample a masking ratio $t\sim\mathcal{U}[0,1]$ and independently mask each token with a probability $t$ to obtain the masked sequence $\hat{\mathbf{y}}_t$.
The resulting objective is
\begin{equation}
\label{eq:dlm}
\begin{aligned}
\mathcal{L}_{\text{DLM}}
&= -\mathbb{E}_{(x,\hat{\mathbf{y}})\sim\mathcal{D}}\,
\mathbb{E}_{t}\Bigg[\\
&\qquad
\frac{1}{t}
\sum_{i=1}^{L_g}
\mathbf{1}\!\big[\hat{y}_{t,i}=\texttt{[MASK]}\big]\,
\log q_\phi\big(\hat{y}_i \mid \hat{\mathbf{y}}_t,x\big)
\Bigg].
\end{aligned}
\end{equation}
Here, $\mathbf{1}[\cdot]$ denotes the indicator function. 

Consequently, the total training objective is
\begin{equation}
\label{eq:total_loss}
\mathcal{L}(\phi)
= w_{\text{distill}}\,\mathcal{L}_{\text{Distillation}}
+ w_{\text{cons}}\,\mathcal{L}_{\text{Consistency}}
+ w_{\text{dlm}}\,\mathcal{L}_{\text{DLM}}.
\end{equation}

\subsection{Inference}
\label{subsec:inference}

At inference time, the student decodes block-wise under a block-causal mask (Figure~\ref{fig:attn-masks}), reusing the KV cache for the prompt and all previously finalized blocks.
Within each block, we apply \emph{confidence-thresholded} parallel finalization: at each step, among the masked positions of the current block, we reveal every token whose confidence exceeds a threshold $\tau_{\text{conf}}$, following Fast-dLLM~\cite{wu2025fastdllm}.
We also adopt early stopping: decoding terminates once an \texttt{<endoftext>} token is produced within the current block, analogous to AR decoding.
We intentionally avoid additional heuristics such as inter-block parallelism~\cite{wang2025d2f}, which introduce extra hyperparameters whose optimal values are task- and domain-dependent.

\begin{table*}[t]
\caption{
Evaluation results for \textbf{Dream-7B-Instruct}. Arrows in headers indicate whether higher ($\uparrow$) or lower ($\downarrow$) is better.
\emph{Notes:} Par.\,= parallel decoding; D.C.\,= dual-cache KV.
}
\label{tab:main-results-dream}
\vspace{1mm}
\centering
\small
\setlength{\tabcolsep}{8pt}
\begin{tabular}{l l c c c c c}
\toprule
\textbf{Benchmark} & \textbf{Method} & \textbf{TPS $\uparrow$} & \textbf{Latency (s) $\downarrow$} & \textbf{Total Steps $\downarrow$} & \textbf{Gen. Length} & \textbf{Score $\uparrow$} \\
\midrule
\specialrule{0.5pt}{0pt}{0pt}
\addlinespace[2pt]

\multirow{5}{*}{\shortstack[l]{\textbf{GSM8K}\\\textbf{-CoT}\\(8-shot)}}
& Dream-7B-Instruct  & 4.1 \multfacblack{1.0} & 23.5 \multfacblack{1.0} & 256.0 \multfacblack{1.0} & 95.1 & 79.1 \\
& dLLM-Cache   & 6.9 \multfacblack{1.7} & 12.6 \multfacblack{1.9} & 256.0 \multfacblack{1.0} & 87.5 & 75.2 \\
& Fast-dLLM (Par.) & 19.2 \multfacblack{4.7} & 5.0 \multfacblack{4.7} & 53.7 \multfacblack{4.8} & 96.0 & \textbf{79.9} \\
& Fast-dLLM (Par.+D.C.) & 36.6 \multfacblack{8.9} & 2.5 \multfacblack{9.4} & 60.8 \multfacblack{4.2} & 90.3 & 77.3 \\
\rowcolor{gray!12}
& CDLM--Dream (ours)           & \textbf{51.7} \multfac{12.6} & \textbf{2.1} \multfac{11.2} & \textbf{44.1} \multfac{5.8} & 107.4 & 78.8 \\
\specialrule{1pt}{0pt}{0pt}
\addlinespace[2pt]

\multirow{5}{*}{\shortstack[l]{\textbf{HumanEval}\\\textbf{-Instruct}\\(0-shot)}}
& Dream-7B-Instruct  & 15.4 \multfacblack{1.0} & 13.4 \multfacblack{1.0} & 256.0 \multfacblack{1.0} & 206.7 & 48.2 \\
& dLLM-Cache   & 9.4 \multfacblack{0.6} & 12.0 \multfacblack{1.1} & 256.0 \multfacblack{1.0} & 113.6 & 43.9 \\
& Fast-dLLM (Par.) & 66.4 \multfacblack{4.3} & 3.2 \multfacblack{4.2} & 61.6 \multfacblack{4.2} & 210.2 & 45.7 \\
& Fast-dLLM (Par.+D.C.) & \textbf{79.9} \multfacblack{5.2} & 2.5 \multfacblack{5.4} & 71.6 \multfacblack{3.6} & 200.9 & 46.3 \\
\rowcolor{gray!12}
& CDLM--Dream (ours)           & 43.3 \multfac{2.8} & \textbf{2.2} \multfac{6.1} & \textbf{49.6} \multfac{5.2} & 96.9 & \textbf{50.0} \\
\specialrule{1pt}{0pt}{0pt}
\addlinespace[2pt]

\multirow{5}{*}{\shortstack[l]{\textbf{MATH}\\(4-shot)}}
& Dream-7B-Instruct  & 8.3 \multfacblack{1.0} & 21.9 \multfacblack{1.0} & 256.0 \multfacblack{1.0} & 182.2 & \textbf{38.0} \\
& dLLM-Cache   & 11.3 \multfacblack{1.4} & 13.0 \multfacblack{1.7} & 256.0 \multfacblack{1.0} & 146.9 & 35.8 \\
& Fast-dLLM (Par.) & 23.9 \multfacblack{2.9} & 7.6 \multfacblack{2.9} & 87.1 \multfacblack{2.9} & 181.0 & 37.8 \\
& Fast-dLLM (Par.+D.C.) & 48.8 \multfacblack{5.9} & 3.7 \multfacblack{5.9} & 98.2 \multfacblack{2.6} & 179.8 & 37.4 \\
\rowcolor{gray!12}
& CDLM--Dream (ours)           & \textbf{53.8} \multfac{6.5} & \textbf{2.9} \multfac{7.6} & \textbf{63.2} \multfac{4.1} & 158.4 & 32.4 \\
\specialrule{1pt}{0pt}{0pt}
\addlinespace[2pt]

\multirow{5}{*}{\shortstack[l]{\textbf{MBPP}\\\textbf{-Instruct}\\(0-shot)}}
& Dream-7B-Instruct  & 2.3 \multfacblack{1.0} & 21.7 \multfacblack{1.0} & 256.0 \multfacblack{1.0} & 49.5 & 51.8 \\
& dLLM-Cache   & 5.1 \multfacblack{2.2} & 11.3 \multfacblack{1.9} & 256.0 \multfacblack{1.0} & 57.3 & \textbf{56.0} \\
& Fast-dLLM (Par.) & 15.5 \multfacblack{6.7} & 3.1 \multfacblack{7.0} & 35.3 \multfacblack{7.3} & 47.3 & 55.6 \\
& Fast-dLLM (Par.+D.C.) & 25.4 \multfacblack{11.0} & 1.9 \multfacblack{11.4} & 43.6 \multfacblack{5.9} & 47.1 & 53.4 \\
\rowcolor{gray!12}
& CDLM--Dream (ours)           & \textbf{48.1} \multfac{20.9} & \textbf{1.5} \multfac{14.5} & \textbf{33.2} \multfac{7.7} & 74.3 & 53.0 \\

\bottomrule
\end{tabular}
\vskip -0.05in
\end{table*}

\begin{table*}[t]
\caption{
Evaluation results for \textbf{LLaDA-8B-Instruct}. Arrows in headers indicate whether higher ($\uparrow$) or lower ($\downarrow$) is better.
We report HumanEval (not HumanEval-Instruct), as the latter lowers overall baseline scores. \emph{Notes:} Par.\,= parallel decoding; D.C.\,= dual-cache KV.
}
\label{tab:main-results-llada}
\vspace{1mm}
\centering
\small
\setlength{\tabcolsep}{8pt}
\begin{tabular}{l l c c c c c}
\toprule
\textbf{Benchmark} & \textbf{Method} & \textbf{TPS $\uparrow$} & \textbf{Latency (s) $\downarrow$} & \textbf{Total Steps $\downarrow$} & \textbf{Gen. Length} & \textbf{Score $\uparrow$} \\
\midrule
\specialrule{0.5pt}{0pt}{0pt}
\addlinespace[2pt]

\multirow{5}{*}{\shortstack[l]{\textbf{GSM8K}\\(4-shot)}}
  & LLaDA-8B-Instruct               & 8.2 \multfacblack{1.0} & 28.3 \multfacblack{1.0} & 256.0 \multfacblack{1.0} & 232.1 & 77.1 \\
  & dLLM-Cache   & 18.7 \multfacblack{2.3} & 12.3 \multfacblack{2.3} & 256.0 \multfacblack{1.0} & 229.6 & 78.5 \\
  & Fast-dLLM (Par.)            & 26.7 \multfacblack{3.3} & 8.7 \multfacblack{3.3} & 77.5 \multfacblack{3.3} & 231.3 & \textbf{78.6} \\
  & Fast-dLLM (Par.+D.C.) & \textbf{54.6} \multfacblack{6.7} & 4.2 \multfacblack{6.7} & 85.0 \multfacblack{3.0} & 230.4 & 76.5 \\
\rowcolor{gray!12}
  & CDLM--LLaDA (ours)              & 54.3 \multfac{6.6} & \textbf{3.3} \multfac{8.6} & \textbf{57.7} \multfac{4.4} & 177.3 & 73.9 \\
\specialrule{1pt}{0pt}{0pt}
\addlinespace[2pt]

\multirow{5}{*}{\shortstack[l]{\textbf{HumanEval}\\(0-shot)}}
  & LLaDA-8B-Instruct            & 7.4 \multfacblack{1.0} & 11.3 \multfacblack{1.0} & 256.0 \multfacblack{1.0} & 83.9 & 37.8 \\
  & dLLM-Cache   & 12.6 \multfacblack{1.7} & 10.8 \multfacblack{1.0} & 256.0 \multfacblack{1.0} & 135.9 & 39.0 \\
  & Fast-dLLM (Par.)     & 17.9 \multfacblack{2.4} & 4.6 \multfacblack{2.5} & 100.3 \multfacblack{2.6} & 83.1 & 36.6 \\
  & Fast-dLLM (Par.+D.C.)   & 18.9 \multfacblack{2.6} & 4.2 \multfacblack{2.7} & 97.7 \multfacblack{2.6} & 80.2 & 36.0 \\
\rowcolor{gray!12}
  & CDLM--LLaDA (ours)           & \textbf{50.9} \multfac{6.9} & \textbf{1.9} \multfac{5.9} & \textbf{32.3} \multfac{7.9} & 95.1 & \textbf{40.2} \\
\specialrule{1pt}{0pt}{0pt}
\addlinespace[2pt]

\multirow{5}{*}{\shortstack[l]{\textbf{MATH}\\(4-shot)}}
  & LLaDA-8B-Instruct            & 8.8 \multfacblack{1.0} & 25.7 \multfacblack{1.0} & 256.0 \multfacblack{1.0} & 226.4 & 24.1 \\
  & dLLM-Cache   & 22.7 \multfacblack{2.6} & 10.9 \multfacblack{2.4} & 256.0 \multfacblack{1.0} & 248.5 & 33.3 \\
  & Fast-dLLM (Par.)     & 25.1 \multfacblack{2.9} & 9.9 \multfacblack{2.6} & 96.9 \multfacblack{2.6} & 248.6 & \textbf{33.4} \\
  & Fast-dLLM (Par.+D.C.)       & 49.7 \multfacblack{5.7} & 5.0 \multfacblack{5.1} & 107.0 \multfacblack{2.4} & 248.2 & 32.5 \\
\rowcolor{gray!12}
  & CDLM--LLaDA (ours)           & \textbf{50.2} \multfac{5.7} & \textbf{4.2} \multfac{6.1} & \textbf{75.3} \multfac{3.4} & 210.3 & 28.3 \\
\specialrule{1pt}{0pt}{0pt}
\addlinespace[2pt]

\multirow{5}{*}{\shortstack[l]{\textbf{MBPP}\\\textbf{-Instruct}\\(0-shot)}}
  & LLaDA-8B-Instruct            & 17.7 \multfacblack{1.0} & 11.4 \multfacblack{1.0} & 256.0 \multfacblack{1.0} & 201.0 & 40.8 \\
  & dLLM-Cache   & 18.5 \multfacblack{1.0} & 10.8 \multfacblack{1.1} & 256.0 \multfacblack{1.0} & 199.4 & 40.8 \\
  & Fast-dLLM (Par.)     & 21.2 \multfacblack{1.2} & 6.2 \multfacblack{1.8} & 59.1 \multfacblack{4.3} & 131.8 & \textbf{42.0} \\
  & Fast-dLLM (Par.+D.C.) & 40.1 \multfacblack{2.3} & 3.4 \multfacblack{3.4} & 66.9 \multfacblack{3.8} & 134.7 & 35.0 \\
\rowcolor{gray!12}
  & CDLM--LLaDA (ours)           & \textbf{60.6} \multfac{3.4} & \textbf{3.2} \multfac{3.6} & \textbf{58.0} \multfac{4.4} & 195.3 & 38.4 \\

\bottomrule
\end{tabular}
\vskip -0.05in
\end{table*}

\section{Experiments}
\label{evaluation}

\subsection{Experimental Setup}
\label{subsec:experimental-setup}

This subsection summarizes datasets, CDLM training, evaluation protocol, baselines, hardware, and metrics.

\vspace{-1mm}
\paragraph{Target Models and Training Datasets}

We consider two representative open-source DLMs: Dream-7B-Instruct~\cite{ye2025dream7b} and LLaDA-8B-Instruct~\cite{nie2025large}.
We derive 7.5k prompts from Bespoke-Stratos-17k~\cite{bespoke_stratos}, filtering to those with prompt length $\leq 512$ tokens.
For LLaDA, we further augment the corpus with 7.5k math-style prompts from DParallel~\cite{chen2025dparallel}, using the same length budget.
When generating teacher trajectories, we fix the generation length to $L_g{=}256$ and the block size to $B{=}32$.
Additional details are provided in Appendix~\ref{appendix:dataset}.

\vspace{-1mm}
\paragraph{Training Details}

We fine-tune all models with LoRA~\cite{hu2022lora} applied to both attention and MLP modules, using a block-wise causal mask (Figure~\ref{fig:attn-masks}) with $B{=}32$ and $L_g{=}256$.
End-to-end training takes approximately \textbf{8 hours} for CDLM--Dream and \textbf{16 hours} for CDLM--LLaDA on $4\times$ NVIDIA A100 (80GB) GPUs.
Additional details are in Appendix~\ref{appendix:training}.

\vspace{-1mm}
\paragraph{Benchmarks.}

Following established conventions~\cite{wang2025d2f, wu2025fastdllm}, we evaluate mathematical reasoning and code generation on GSM8K, GSM8K-CoT~\cite{cobbe2021gsm8k}, MATH~\cite{hendrycks2021math}, HumanEval~\cite{chen2021humaneval}, and MBPP~\cite{austin2021mbpp}.
Coding tasks are evaluated zero-shot, and few-shot settings for math follow the original Dream and LLaDA papers.
All efficiency measurements are taken on $4\times$ NVIDIA A100\,(80\,GB) GPUs with batch size $1$ under data parallelism.
Latency, total steps, and generation length are reported as per-sample averages over the evaluation set.
Additional evaluation details are provided in Appendix~\ref{appendix:evaluation}.

\paragraph{Baselines.}

As references, we evaluate the original DLMs using block-wise decoding under their official inference settings.
Because our work targets two orthogonal axes of inference acceleration, (1) \emph{KV caching} and (2) \emph{step reduction}, we include three inference-time baselines:
(i) dLLM-Cache~\cite{liu2025dllmcache}, which focuses on (1) via adaptive feature caching;
(ii) Fast-dLLM (Parallel)~\cite{wu2025fastdllm}, which targets (2) via confidence thresholding; and
(iii) Fast-dLLM (Parallel + Dual Cache)~\cite{wu2025fastdllm}, which combines (1)\,+\,(2) via approximate dual-cache KV caching.
We omit D2F~\cite{wang2025d2f} here because it is trained for $L_g{=}512$ and reports on Dream-7B-Base rather than Dream-7B-Instruct.
For autoregressive comparisons, we evaluate Qwen2.5-7B-Instruct~\cite{qwen2024techreport} alongside Dream and Llama-3.1-8B-Instruct~\cite{grattafiori2024llama} alongside LLaDA.
For fair comparison across methods, we fix $L_g{=}256$, use greedy decoding (temperature $0.0$), apply the same $B{=}32$, and set the token-confidence threshold to $\tau_{\text{conf}}{=}0.9$.
Further details are in Appendix~\ref{appendix:evaluation}.

\subsection{Main Results}
\label{subsec:main-results}

We compare CDLM against vanilla DLMs, accelerated DLM variants, and autoregressive baselines.

\subsubsection{Results on CDLM--Dream}
Table~\ref{tab:main-results-dream} summarizes throughput, latency, step counts, generation length, and accuracy for Dream-7B-Instruct and its acceleration baselines.

\paragraph{Steps and scores.}
Our objective is to reduce the number of sampling steps so that the model reaches a coherent answer more quickly.
As shown in Table~\ref{tab:main-results-dream}, CDLM--Dream achieves the largest step reductions across benchmarks, cutting refinement steps by roughly $4.1\times$--$7.7\times$ with minor accuracy changes on most benchmarks. Naive Dream is run with 256 refinement steps at its most performant setting; simply reducing the step count markedly degrades generation quality (see Section~\ref{subsubsec:steps}).
Despite aggressive step reduction, CDLM--Dream matches or improves accuracy: it rises from 51.8$\to$53.0 on MBPP-Instruct and from 48.2$\to$50.0 on HumanEval-Instruct, with only a negligible drop on GSM8K-CoT (79.1$\to$78.8).
In contrast, dLLM-Cache keeps the step budget fixed at 256 and accelerates via KV caching rather than step reduction.

We attribute the accuracy drop on MATH to two factors.
First, the small training mixture ($\sim$7.5k prompts) provides limited exposure to the advanced reasoning patterns required by MATH. 
The data are drawn from filtered subsets that largely contain problems already solvable by Qwen2.5, resulting in a narrower effective difficulty range.
Second, teacher trajectories are capped at 256 generation tokens during training.
This shorter reasoning budget may be inadequate for complex multi-step problems.
A natural next step is to train with a longer budget (e.g., 512 tokens), as in D2F~\cite{wang2025d2f}.

\paragraph{Latency and Throughput (TPS).}
CDLM--Dream shows the largest latency reductions across all benchmarks, with speedups of up to $11.2\times$ on GSM8K-CoT and $14.5\times$ on MBPP-Instruct.
These gains stem from combining \emph{step reduction} with \emph{block-wise KV caching}.
The impact of caching is evident when comparing methods with similar refinement steps.
For example, on GSM8K-CoT, Fast-dLLM (Par.+D.C.) halves latency (5.0\,s$\to$2.5\,s) relative to Fast-dLLM (Par.) despite using more refinement steps (53.7$\to$60.8) by avoiding redundant computation.
CDLM--Dream further benefits from block-wise causal training (Figure~\ref{fig:attn-masks}), which enables exact KV caching and early termination at block boundaries, reducing unnecessary decoding.

CDLM--Dream attains the highest TPS on three out of four benchmarks.
The sole exception is HumanEval-Instruct, where TPS is lower than Fast-dLLM baselines due to a substantially shorter effective generation length (97 vs.\ $\sim$200).
Differences in decoding dynamics can arise because CDLM is strictly block-causal and cannot attend to future blocks, lacking explicit budget awareness, whereas other baselines retain full bidirectional attention that preserves a form of global budget awareness.
Despite shorter generations, CDLM--Dream achieves higher pass@1, indicating that it can emit fewer tokens while preserving solution quality.

\subsubsection{Results on CDLM--LLaDA}
Table~\ref{tab:main-results-llada} summarizes throughput, latency, step counts, generation length, and accuracy for LLaDA-8B-Instruct and its acceleration baselines. 

\paragraph{Steps and scores.}

As shown in Table~\ref{tab:main-results-llada}, CDLM--LLaDA delivers the largest step reductions across all four benchmarks, cutting steps by roughly $3.4\times$--7.9$\times$. 
Despite this aggressive reduction, CDLM--LLaDA improves HumanEval from 37.8$\to$40.2 and MATH from 24.1$\to$28.3. 
MBPP-Instruct shows a slight drop (40.8$\to$38.4), yet still outperforms Fast-dLLM (Par.+D.C.) by $3.4$ points at comparable throughput. 
GSM8K is the only benchmark with a clear degradation (77.1$\to$73.9), which we analyze below.

We attribute the GSM8K degradation primarily to limited data and LLaDA’s sensitivity to fine-tuning.
In initial experiments, training CDLM--LLaDA on bespoke $\sim$7.5k prompts already reduced GSM8K accuracy (77.1$\to$72.1), consistent with prior observations that even simple SFT can harm math performance unless data are carefully curated and sufficiently large.
Augmenting the data with an additional 7.5k math-focused prompts partially recovers performance (GSM8K 72.1$\to$73.9, MATH 25.4$\to$28.3) while preserving coding scores.
In addition, the fixed generation budget ($L_g{=}256$) may further constrain reasoning depth, as LLaDA tends to produce longer outputs when allowed larger budgets.
We expect that jointly scaling the dataset and increasing the generation budget would close the remaining GSM8K gap.

\paragraph{Latency and Throughput (TPS).}

CDLM--LLaDA attains the strongest latency reductions across all tasks; for example, on GSM8K latency falls from 28.3\,s$\to$3.3\,s.
Throughput gains are likewise substantial, ranging from $3.4\times$ to $6.9\times$ relative to the naive LLaDA baseline.
The role of KV caching is evident when step counts are similar: LLaDA vs.\ dLLM-Cache keeps $256$ steps yet nearly halves latency on GSM8K (28.3\,s$\to$12.3\,s), and Fast-dLLM (Par.) vs.\ (Par.+D.C.) reduces GSM8K latency further (8.7\,s$\to$4.2\,s) even as steps increase (77.5$\to$85.0).
CDLM--LLaDA goes beyond caching alone by also reducing refinement steps via \emph{consistency} training while retaining block-wise causality for exact KV caching, yielding the best overall efficiency in both latency and TPS.

\begin{figure}[t]
  \centering
  \includegraphics[width=0.98\linewidth]{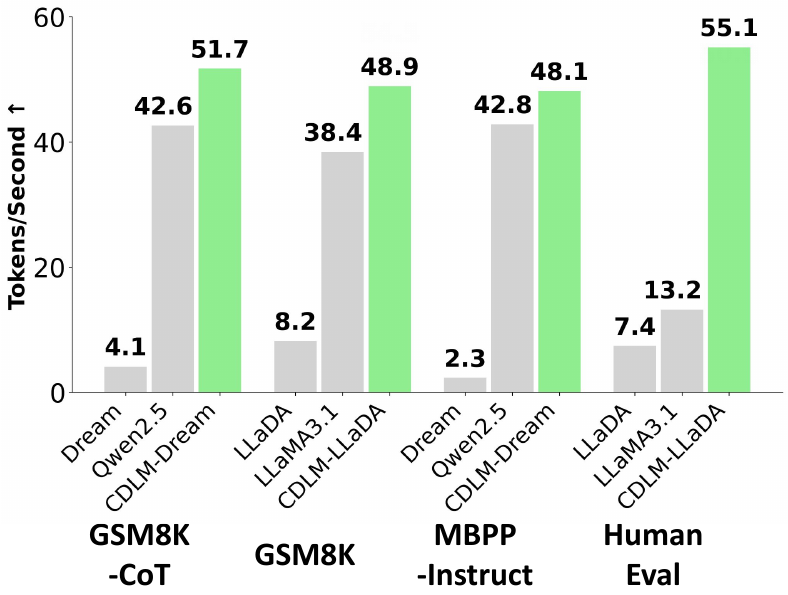}
  \caption{
  \textbf{Throughput comparison across benchmarks.}
  Tokens per second on GSM8K, MBPP, and HumanEval for Dream-7B-Instruct and LLaDA-8B-Instruct under naive diffusion decoding, autoregressive (AR) baselines, and CDLM.
  }
  \label{fig:ar-tps-comparison}
\end{figure}

\subsubsection{Comparison with Autoregressive Models}

\paragraph{Throughput.}
CDLM improves naive DLMs in throughput by \(3\times\)--\(21\times\) and also surpasses equal-size autoregressive (AR) baselines (Figure~\ref{fig:ar-tps-comparison}).
CDLM--Dream achieves \(1.2\times\) and \(1.1\times\) higher throughput than its AR baseline on GSM8K-CoT (8-shot) and MBPP-Instruct (0-shot), respectively, while CDLM--LLaDA attains \(1.3\times\) and \(4.2\times\) higher throughput on GSM8K (4-shot) and HumanEval (0-shot).
Throughput here reflects both per-step cost and the number of refinement steps.
A CDLM refinement step is more expensive than an AR step due to matrix-matrix multiplications for decoding. However, CDLM can finalize multiple tokens per iteration, reducing total steps and increasing effective throughput.
For example, from Table~\ref{tab:main-results-dream}, CDLM--Dream produces on average \(107.4/44.1 \approx 2.4\) tokens per step on GSM8K-CoT and \(74.3/33.2 \approx 2.2\) tokens per step on MBPP-Instruct.

\paragraph{Accuracy.}
Although throughput is our focus, accuracy may be lower than that of AR models because CDLMs remain bounded by the strength of their DLM backbones.
For CDLM--Dream versus its AR baseline, the accuracies are \(78.8\) vs.\ \(73.8\) on GSM8K-CoT and \(53.0\) vs.\ \(81.7\) on MBPP-Instruct (CDLM vs.\ AR).
For CDLM--LLaDA versus its AR baseline, the respective accuracies are \(73.9\) vs.\ \(80.3\) on GSM8K and \(40.2\) vs.\ \(60.4\) on HumanEval (CDLM vs.\ AR).
As stronger DLM backbones emerge, applying our consistency-based fine-tuning should yield further throughput gains with improved quality.

\subsection{Ablation Studies}

We analyze how training and inference choices affect efficiency and accuracy.

\begin{table}[t]
\caption{\textbf{Ablation of loss weights.}
Effects of varying $(w_{\text{distill}}, w_{\text{cons}}, w_{\text{dlm}})$ on \emph{score} (black) and \emph{steps to convergence} (blue) on GSM8K and HumanEval-Instruct for CDLM--Dream.}
\label{tab:ablation-loss-coeff}
\vspace{1mm}
\centering
\small
\renewcommand{\arraystretch}{1.2}

{\setlength{\tabcolsep}{4pt}
 \rowcolors{2}{gray!10}{white}
\begin{tabular}{@{} c c c C{2.2cm} C{2.4cm} @{}}
\toprule
$\boldsymbol{w_{\text{distill}}}$ & $\boldsymbol{w_{\text{cons}}}$ & $\boldsymbol{w_{\text{dlm}}}$ &
\shortstack{\textbf{GSM8K}\\ \small(4-shot)} &
\shortstack{\textbf{HumanEval-Inst.}\\ \small(0-shot)} \\
\midrule
1.0 & \ballotx & 0.01 & \metriccell{73.2}{46.7} & \metriccell{42.7}{61.0} \\
\ballotx & 1.0 & 0.01 & \metriccell{6.9}{100.6} & \metriccell{0.0}{124.3} \\
1.0 & 1.0 & 0.01 & \metriccell{74.1}{49.4} & \metriccell{42.7}{49.4} \\
1.0 & 1.0 & \ballotx & \metriccell{73.3}{46.2} & \metriccell{48.2}{51.5} \\
1.0 & 0.1 & 0.01 & \metriccell{75.1}{48.0} & \metriccell{45.7}{59.9} \\
1.0 & 0.1 & \ballotx & \metriccell{73.7}{48.1} & \metriccell{50.6}{69.0} \\
\bottomrule
\end{tabular}
}
\vskip -0.05in
\end{table}

\subsubsection{Loss-Weight Composition}
\label{subsubsec:loss}

Table~\ref{tab:ablation-loss-coeff} varies the loss weights \((w_{\text{distill}}, w_{\text{cons}}, w_{\text{dlm}})\) to study the trade-off between convergence speed and final accuracy.
We train the Dream model for four epochs with a constant learning rate \texttt{2e-5}, then evaluate on GSM8K (4-shot) and HumanEval-Instruct (0-shot).

\paragraph{Distillation vs.\ consistency in isolation.}

With a small auxiliary DLM loss ($w_{\text{dlm}}{=}0.01$), the \emph{distillation-only} setting (row~1) acts as a strong anchor: it converges quickly (about 46.7 and 61.0 steps for a generation length of 256), with a slight degradation in score.
In contrast, \emph{consistency-only} (row~2) collapses: optimizing for self-agreement without teacher supervision causes failure.

\paragraph{Coupling yields synergy.}

When we couple \emph{consistency} with \emph{distillation} (rows~3 and~5), convergence becomes both faster and more stable.
GSM8K improves from 73.2 to 74.1--75.1 with similar or fewer steps, and HumanEval is comparable to or higher (42.7$\to$45.7) while requiring fewer iterations.
We fix \(w_{\text{distill}}{=}1.0\) and set \(w_{\text{cons}}{=}0.5\) as a good balance between speed and quality.

\paragraph{Role of the auxiliary DLM loss.}

Removing the ground-truth masked-denoising term (rows~4 and~6) raises HumanEval but lowers GSM8K, indicating a loss of math reasoning ability.
A small auxiliary weight helps preserve this ability, and we find that general coding performance can be recovered with longer training.
Accordingly, we use \(w_{\text{dlm}}{=}0.01\) for Dream and \(w_{\text{dlm}}{=}0.1\) for LLaDA, as the DLM loss on LLaDA has a smaller absolute scale.

\subsubsection{Effective Step Reduction}
\label{subsubsec:steps}

We evaluate a naive step-truncation baseline by forcing the teacher DLMs to use a similar number of refinement steps as our CDLMs (Table~\ref{tab:ablation-naive-steps}).
Because decoding is block-wise, the step count must be a multiple of the block factor ($L_g/B{=}8$).
For Dream, we set the step budget to $48$ and measure GSM8K-CoT (8-shot) as in Table~\ref{tab:main-results-dream}; for LLaDA, we set it to $56$ and measure GSM8K (4-shot) as in Table~\ref{tab:main-results-llada}.
Naively truncating the step budget, in other words, forcing a non-retrained DLM to finalize multiple tokens per iteration, leads to marked accuracy degradation.
By contrast, CDLM maintains quality at comparable step counts, underscoring that consistency training is necessary for stable multi-token refinement. Furthermore, with KV caching, our approach reduces latency by roughly half relative to the naive DLM at comparable iteration counts.

\begin{table}[t]
\caption{\textbf{Ablation of refinement steps.}
GSM8K results obtained by naively truncating the baseline DLM step counts
to match the step budgets used by CDLM--Dream and CDLM--LLaDA.}
\label{tab:ablation-naive-steps}
\vspace{1mm}
\centering
\small
\setlength{\tabcolsep}{6pt}
\begin{tabular}{lccc}
\toprule
\textbf{Method} &  \textbf{Latency (s) $\downarrow$} & \textbf{Steps $\downarrow$} & \textbf{Score $\uparrow$} \\
\midrule
Dream-7B-Instruct & 4.4 & 48 & 41.8 \\
CDLM--Dream (ours)    & 2.1 & 44.1 & 78.8 \\
\specialrule{1pt}{0pt}{0pt}
\addlinespace[2pt]
LLaDA-8B-Instruct & 6.0 & 56 & 60.3 \\
CDLM--LLaDA (ours)    & 3.3 & 57.7 & 73.9 \\
\bottomrule
\end{tabular}
\vskip -0.05in
\vspace{-0.1in}
\end{table}

\begin{figure}[t]
  \centering
  \vspace{2mm}
  \includegraphics[width=0.98\linewidth]{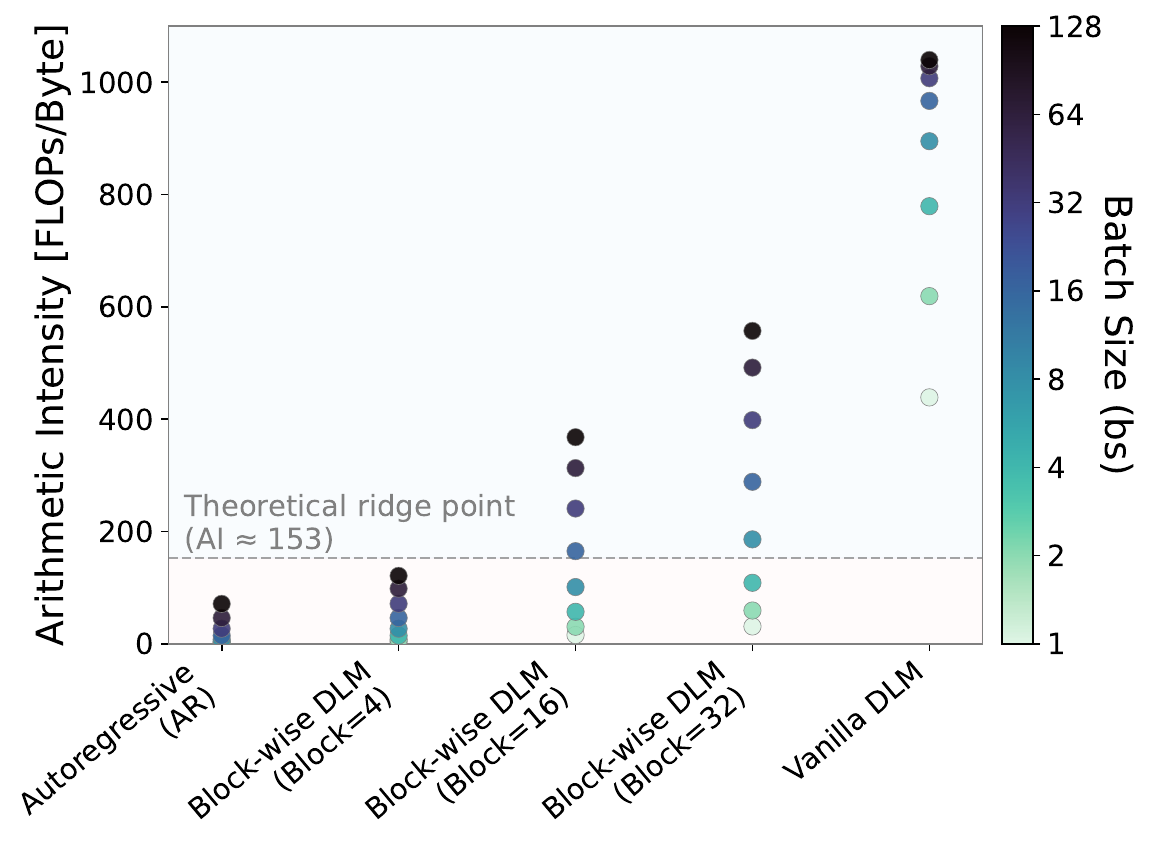}
  \vspace{-5mm}
  \caption{
  \textbf{Arithmetic intensity across batch sizes.}
  Arithmetic Intensity (AI) during decoding as a function of batch size ($\mathrm{bs}\in\{1,2,4,8,16,32,128\}$) for autoregressive (AR) models, vanilla DLMs, and block-wise DLMs (CDLM).
  The dashed line indicates the theoretical ridge point, separating memory-bound (red) and compute-bound (blue) regimes.}
  \label{fig:ai-analysis}
\end{figure}

\subsection{System-Level Scalability via Arithmetic Intensity}
\label{subsubsec:system-analysis}

To understand CDLM’s hardware-utilization characteristics, we analytically model the inference for autoregressive (AR) models, vanilla DLMs, and block-wise DLMs with block sizes $B \in \{4,16,32\}$.
Building on prior modeling frameworks~\cite{tiwari2025quantspec, kim2025beyond}, we parameterize the AR baseline following the LLaMA-3.1~\cite{grattafiori2024llama} configuration, while vanilla and block-wise DLMs follow the LLaDA~\cite{nie2025large} configuration.
We examine \emph{arithmetic intensity} (AI), the ratio of computation to memory traffic (FLOPs per byte moved), to diagnose whether inference is memory- or compute-bound and to characterize batch-level scalability.
Concretely, we consider a prompt length $L_p{=}512$ and generation length $L_g{=}256$ to match Section~\ref{subsec:main-results}, and track AI as the batch size ($\mathrm{bs}$) increases from 1 to 128 (Figure~\ref{fig:ai-analysis}).
To focus on decoding behavior, we exclude one-time prefill costs (KV-cache initialization) and ignore memory-capacity constraints (e.g., out-of-memory).

\paragraph{Autoregressive (AR) Models.}
AR decoding operates in a strongly memory-bound regime, with AI close to 1 at $\mathrm{bs}=1$.
This is because each decoding step processes a single token with substantial weight/KV-cache traffic.
As the batch size increases, weight traffic is amortized across sequences, resulting in approximately linear AI scaling at small batch sizes (1.0$\to$2.0$\to$4.0$\to$7.8 for $\mathrm{bs}\in\{1,2,4,8\}$).
However, even at $\mathrm{bs}=128$ (AI 71.3), AR decoding remains below the compute-bound regime, indicating that performance is still primarily constrained by memory traffic.

\paragraph{Vanilla DLMs.}
In contrast, vanilla DLMs are compute-bound even at $\mathrm{bs}=1$, with AI (438.9) already exceeding the ridge point.
Each denoising step recomputes the entire sequence with full bidirectional attention, rather than reusing a KV cache across steps, leading to a highly compute-intensive workload.
As a result, increasing $\mathrm{bs}$ yields limited additional AI scaling (438.9$\to$619.2$\to$779.3 for $\mathrm{bs}\in\{1,2,4\}$), and AI nearly saturates beyond $\mathrm{bs}=64$ (1028.6$\to$1039.7 from 64 to 128).

\paragraph{Block-wise DLMs (CDLM).}
Block-wise DLMs operate in an intermediate regime.
At $\mathrm{bs}=1$, their AI (4.0, 15.8, 31.1 for $B\in\{4,16,32\}$) exceeds AR decoding but remains below vanilla DLMs.
This gain stems from intra-block parallelism: each step processes $B$ tokens under comparable memory traffic as a single-token AR step, effectively amortizing weight loads across the block and scaling AI by nearly $B$.
The AI crosses the ridge point at relatively small batch sizes (e.g., $B{=}32$ at $\mathrm{bs}\approx8$ and $B{=}16$ at $\mathrm{bs}\approx16$), allowing block-wise DLMs to better utilize available compute in small-batch settings.
At higher batch sizes, however, the marginal AI gain diminishes as $\mathrm{bs}$ increases, since block-wise DLMs already exploit both batch- and block-level parallelism.

Overall, this analysis shows that CDLM-like block-wise DLMs sit between the memory-bound regime of AR models and the compute-saturated regime of full-attention DLMs.
This balanced operating point explains CDLM’s superior throughput and scalability in small-batch inference.
A detailed roofline analysis with AI and corresponding simulated performance is provided in Appendix~\ref{appendix:roofline-analysis}.

\section{Discussion}
\label{discussion}
Here, we discuss key trade-offs, potential extensions, and limitations of our approach.

\paragraph{Expressiveness vs. efficiency trade-off}

Full bidirectional attention in DLMs requires recomputing $\mathcal{O}(L^2)$ attention at every denoising step, making inference highly compute-intensive and impractical at scale.
CDLM enables exact KV caching to mitigate this cost while preserving bidirectional context within each block, retaining local refinement capabilities such as infilling inside the current block.
We further believe that the mild left-to-right inductive bias introduced by block-wise decoding aligns well with reasoning-style generation.

\paragraph{Inference-only acceleration techniques}

Our method is orthogonal to and can be combined with various inference-only (training-free) techniques.
At inference time, we only use block-wise KV caching and confidence-based multi-token finalization, but more sophisticated mechanisms can be layered on top.
For example, incorporating D2F~\cite{wang2025d2f}’s inter-block parallelism could further reduce wall-clock latency.

\paragraph{Scaling with stronger DLM backbones}

CDLM is a post-training recipe that can be applied to any block-diffusion models to accelerate convergence, and its effectiveness is expected to grow as stronger DLMs become available.
Recent models such as SDAR~\cite{cheng2025sdar} are reported to outperform Dream and LLaDA and have been released at multiple scales.
A promising direction is to generate trajectories from large-scale DLM teachers (e.g., 30B) and train mid-scale students (e.g., 8B) using the CDLM post-training recipe.

\paragraph{Limitations.}

Training currently relies on offline, static trajectories that are predominantly math-focused.
Despite careful dataset selection, the student may overfit to the teacher’s prior and domain, potentially limiting generalization beyond the training distribution.
For LLaDA, we observe that augmenting the training set with additional math-style prompts improves math performance by 2--3 percentage points while preserving coding ability, suggesting that scaling and diversifying the corpus are straightforward directions for future work.
Furthermore, CDLM’s performance is ultimately bounded by the teacher: a bidirectional DLM distilled into a block-causal student cannot exceed the teacher’s knowledge.
Distilling from stronger DLMs or autoregressive teachers is a natural way to lift this ceiling.
Additional discussion appears in Appendix~\ref{appendix:future}.

\section{Conclusion}
\label{conclusion}

We presented \textbf{CDLM}, a training-based acceleration scheme that brings \emph{consistency modeling} to DLMs.
By enforcing within-block temporal consistency and fine-tuning a block-wise causal student, CDLM reduces refinement steps and enables exact KV caching.
Across math and coding tasks, it yields faster inference, fewer steps, lower latency, and higher throughput while maintaining competitive accuracy.
Promising directions include scaling the distillation corpus, broadening domain coverage, and distilling from stronger teachers.

\section*{Acknowledgements}

We acknowledge gracious support from the FuriosaAI, Intel, Apple, NVIDIA, Macronix, and Mozilla team.
Furthermore, we appreciate support from
Google Cloud, the Google TRC team and Prof. David Patterson.
Prof. Keutzer's lab is sponsored by the Intel corporation, UC Berkeley oneAPI Center of Excellence, Intel VLAB team, as well as funding through BDD and BAIR.
We also acknowledge support by the Director, Office of Science, Office of Advanced Scientific Computing Research, of the U.S. Department of Energy under Contract No. DE-AC02-05CH11231.
DOE SciGPT grant.
Minseo also gratefully acknowledges the SNUCSE GPU Service 3.0 program and Bacchus for providing A100 GPU resources that supported this research.
Our conclusions do not necessarily reflect the position or the policy of our sponsors, and no official endorsement should be~inferred.


\bibliography{references}
\bibliographystyle{mlsys2025}


\appendix

\begin{figure*}[t]
  \centering
  \includegraphics[width=\textwidth]{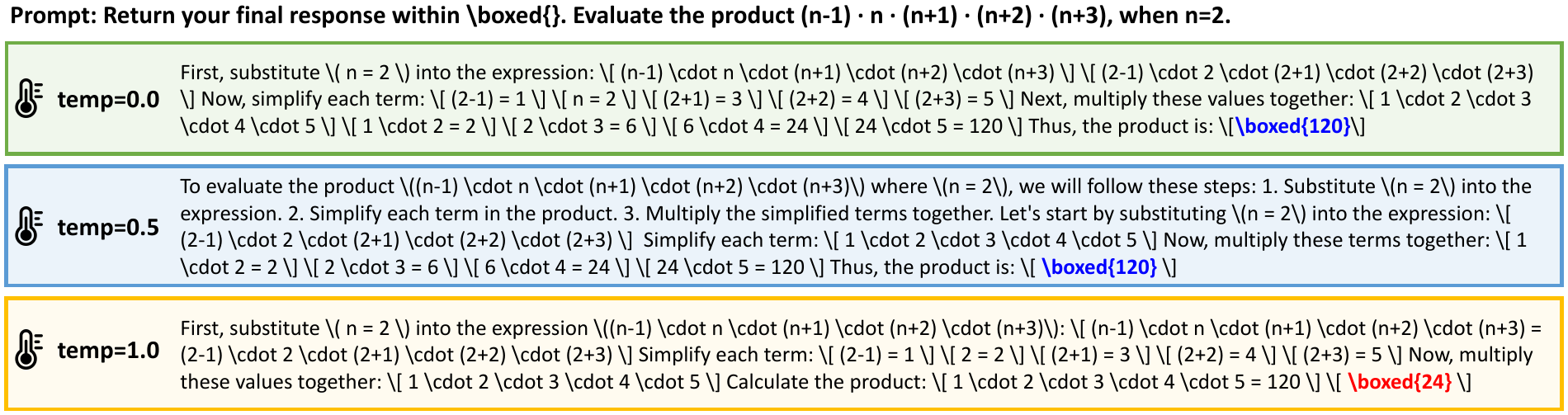}
  \vspace{-7mm}
  \caption{\textbf{Teacher outputs vs.\ temperature.}
  Final outputs from LLaDA-8B-Instruct at sampling temperatures $\tau \in \{0.0,\,0.5,\,1.0\}$.
  Answers are marked with boxed\{$\cdot$\} (\textcolor{blue}{blue} = correct; \textcolor{red}{red} = incorrect).
  }
  \label{fig:appendix-temperature}
\end{figure*}

\begin{figure*}[t]
  \centering
  \includegraphics[width=\textwidth]{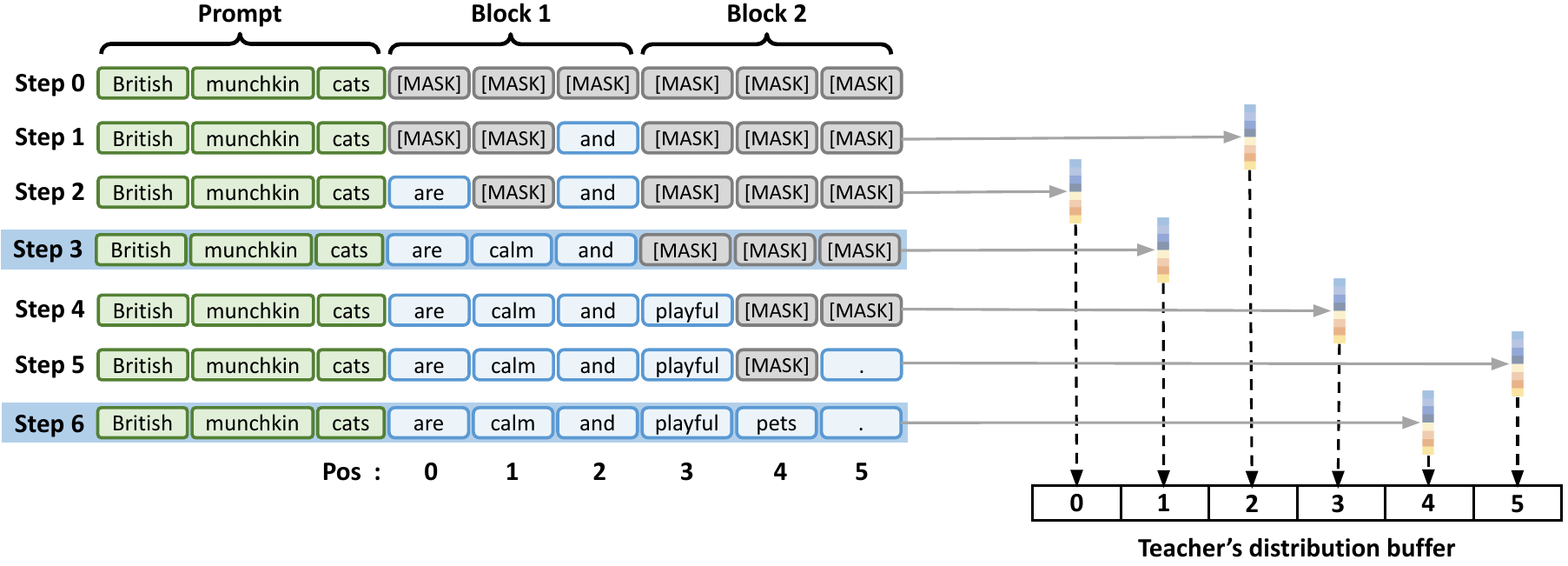}
  \vspace{-7mm}
  \caption{
  \textbf{Teacher trajectory and hidden-state buffer.}
    \textbf{Left:} Block-wise decoding trajectory. 
    \textbf{Right:} As each token is finalized, the teacher’s last hidden state at that position is written to a fixed-length buffer of size $L_g$; arrows indicate write indices across steps (toy example with $B{=}3$, $L_g{=}6$).
  }
  \label{fig:appendix-buffer}
\end{figure*}

\section{Additional Experimental Details}

\subsection{Dataset and Preprocessing}
\label{appendix:dataset}

\paragraph{Sources.}
We use two publicly available datasets from the Hugging Face Hub~\cite{lansechen_easy_2025,lansechen_hard_2025}, which are post-processed versions of Bespoke-Stratos 17k~\cite{bespoke_stratos}.
We choose them because (i) D2F~\cite{wang2025d2f}, a representative
fine-tuning-based DLM acceleration method, uses the same source, and
(ii) they are filtered to a maximum length of 600 tokens and predominantly
contain math word-style reasoning problems.
For both Dream and LLaDA, we treat Qwen2.5-7B’s responses as ground-truth
answers.

Dream is trained only on this Bespoke-derived subset.
To strengthen math performance for LLaDA, we further augment its corpus with an additional $7.5$k math-style prompts sampled from the DParallel
dataset~\cite{chen2025dparallel}.
The released DParallel corpus does not include Qwen2.5-7B outputs, so we
generate reference answers with Qwen2.5-7B for these prompts, yielding an
LLaDA training set whose size is approximately $2\times$ that of Dream.
After trying multiple datasets, we empirically observe that dataset
selection is crucial; overly rigid formats (e.g., multiple-choice math) tend to hurt the model’s ability to generalize.

\paragraph{Preprocessing and Generation Configuration.}
We retain prompts whose input length is at most $512$ tokens.
For each prompt, we generate teacher trajectories using the bidirectional, block-wise DLM, with generation length $L_g{=}256$, block size $B{=}32$, total diffusion steps $N{=}256$, and prompts left-padded to a length of 512 tokens.
Ground-truth responses are truncated or padded to a length of 256 tokens.

\paragraph{Temperature augmentation.}
Effective training benefits from trajectories that are both diverse and reliable.
In DLMs, temperature influences not only token choices but also the \emph{order} in which tokens are revealed~\cite{gong2025diffucoder}, enabling additional diversity in the generated trajectories.
We therefore generate multiple trajectories per prompt at different temperatures.
As shown in Figure~\ref{fig:appendix-temperature}, using temperature $\tau{=}1.0$ can destabilize the reasoning chain and lead to incorrect conclusions.
Consequently, we restrict trajectory collection to $\tau\in\{0.0,\,0.5\}$.

\paragraph{Hidden-state buffer.}
Beyond token trajectories, we store the teacher’s decisions in a compact hidden-state buffer.
Whenever a token is finalized, we record the teacher’s last-layer hidden state at that position; over a generation of length $L_g$, this yields a buffer of shape $d\times L_g$ per example (see Figure~\ref{fig:appendix-buffer}).
Directly saving logits is impractical because their dimensionality is $|V|$ (vocabulary size); using hidden states of dimension $d$ instead gives roughly a $30\times$ reduction in storage.
The trajectory datasets fit on a single server: each shard is \(25\text{-}30\,\mathrm{GiB}\) for \(15\mathrm{k}\) samples.

\subsection{Training}
\label{appendix:training}

\begin{table}[t]
\caption{Training configuration for \textbf{CDLM--Dream-7B-Instruct}.}
\label{tab:cfg-dream}
\centering
\footnotesize
\setlength{\tabcolsep}{3pt}
\renewcommand{\arraystretch}{1.06}
\begin{tabular}{@{} l V @{}}
\toprule
Learning rate & $2\times10^{-5}$ \\
Scheduler (warmup) & constant (5\%) \\
Epochs (best) & 16 (best: 12) \\
Effective batch size & 64 \\
Optimizer & AdamW \\
LoRA rank / $\alpha$ & 32 / 32 \\
LoRA targets &
\makecell[l]{\texttt{q\_proj, k\_proj, v\_proj,}\\
            \texttt{o\_proj, up\_proj,}\\
            \texttt{down\_proj, gate\_proj}} \\
Prompt / generation length & 512 / 256 \\
Loss weights $(w_{\text{distill}}, w_{\text{cons}}, w_{\text{dlm}})$ & (1.0,\; 0.5,\; 0.01) \\
\bottomrule
\end{tabular}
\vspace{-2mm}
\end{table}

\begin{table}[t]
\caption{Training configuration for \textbf{CDLM--LLaDA-8B-Instruct}.}
\label{tab:cfg-llada}
\centering
\footnotesize
\setlength{\tabcolsep}{3pt}
\renewcommand{\arraystretch}{1.06}
\begin{tabular}{@{} l V @{}}
\toprule
Learning rate & $1\times10^{-5}$ \\
Scheduler (warmup) & constant (5\%) \\
Epochs (best) & 16 (best: 12) \\
Effective batch size & 64 \\
Optimizer & AdamW \\
LoRA rank / $\alpha$ & 64 / 64 \\
LoRA targets &
\makecell[l]{\texttt{q\_proj, k\_proj, v\_proj,}\\
            \texttt{attn\_out, up\_proj,}\\
            \texttt{ff\_proj, ff\_out}} \\
Prompt / generation length & 512 / 256 \\
Loss weights $(w_{\text{distill}}, w_{\text{cons}}, w_{\text{dlm}})$ & (1.0,\; 0.5, \; 0.1) \\
\bottomrule
\end{tabular}
\vspace{-2mm}
\end{table}

\paragraph{Training Configurations.}

Models are optimized with AdamW using a constant learning-rate schedule with warmup.
Prompts are left-padded to $512$ tokens and decoded with generation length $256$, resulting in a fixed sequence length of $768$.
Based on the ablation study in Section~\ref{subsubsec:loss}, we set the loss weights $(w_{\text{distill}}, w_{\text{cons}}, w_{\text{dlm}})$ to $(1.0,\,0.5,\,0.01)$ for Dream and $(1.0,\,0.5,\,0.1)$ for LLaDA.
All models are trained for $16$ epochs with an effective batch size of $64$, and we select the best checkpoint based on validation performance.
Training ran on $4\times$ NVIDIA A100 (80\,GB) or $8\times$ NVIDIA RTX A6000 (48\,GB) GPUs; each epoch took roughly $30$ minutes for $15$k trajectories.
We used per-device batch sizes of $1$--$2$ with gradient accumulation to keep the effective batch size fixed at $64$.
Full configurations are listed in Tables~\ref{tab:cfg-dream} and~\ref{tab:cfg-llada}.

\paragraph{Loss formulations.}
Empirically, the \emph{forward} KL divergence yielded more stable, better-calibrated training dynamics and more monotonic convergence than the \emph{reverse} KL divergence for distribution matching.
We also found that distillation in \emph{logit space} outperformed embedding-space distillation using mean squared error (MSE).

\subsection{Evaluation}
\label{appendix:evaluation}

\paragraph{Hardware and parallelism.}

All measurements in Section~\ref{evaluation} except Table~\ref{tab:ablation-loss-coeff} were taken on 4$\times$ NVIDIA A100-80\,GiB GPUs using data parallelism with batch size~$1$ per GPU.

\paragraph{Tooling and post-processing.}

We use the \texttt{lm\_eval} framework (EleutherAI’s \texttt{lm-eval-harness}) for dataset loading, prompting, and evaluation.
We apply the harness’s default post-processing rules, truncating decoded outputs at any task-specific \texttt{stop\_sequence}s prior to scoring.
Evaluation metrics include exact match scores on GSM8K and math-verify scores on MATH.
For code-generation benchmarks (HumanEval and MBPP), we additionally employ Python
post-processing scripts adapted from the DParallel repository~\cite{chen2025dparallel}
to execute generated code and compute pass@1.

\paragraph{Timing and metrics.}

We report \emph{per-sample averages}. 
End-to-end latency is computed by summing the wall-clock time of the generation routine across all ranks and dividing by the total number of evaluated samples.
\emph{Total steps} denotes the average number of refinement steps actually executed per sample (summed across ranks, then divided by the sample count).
\emph{Generation length} is the average number of valid tokens produced per sample, excluding \texttt{<endoftext>} and any content following a task-specific \texttt{stop\_sequence}.

\paragraph{Prompt formatting.}

We follow each benchmark’s standard few-shot protocol and, by default, apply the model’s chat template.
For HumanEval with LLaDA-8B-Instruct, however, enabling the chat template caused a sharp pass@1 decline (naive LLaDA: 37.8$\to$7.9), so we evaluate HumanEval without the chat template.
For math benchmarks, we do not use \texttt{--fewshot\_as\_multiturn}. Instead, all few-shot examples are merged into a single user prompt.

\paragraph{Inference settings.}

For LLaDA, we enable block-wise decoding with block size $B{=}32$.
Dream does not natively support block-wise decoding, so we extend the baseline implementation to support block-wise decoding under the same configuration.

\section{Additional Results}
\label{appendix:additional-results}

\begin{figure*}[t]
  \centering
  \includegraphics[width=\textwidth]{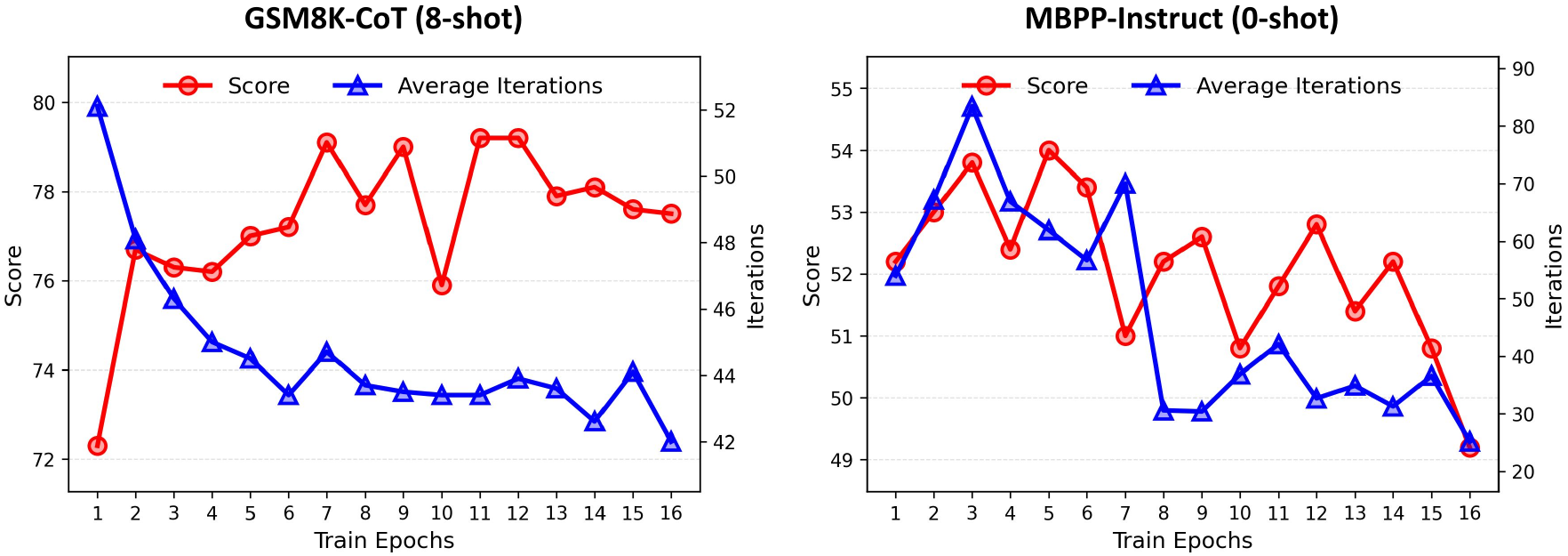}
  \vspace{-7mm}
  \caption{\textbf{Validation trends during CDLM--Dream training.}
   Validation score (red) and average refinement iterations (blue) across training epochs on GSM8K-CoT (8-shot, left) and MBPP-Instruct (0-shot, right).
  }
  \label{fig:appendix-validation}
\end{figure*}

\begin{figure*}[t]
  \centering
  \includegraphics[width=\textwidth]{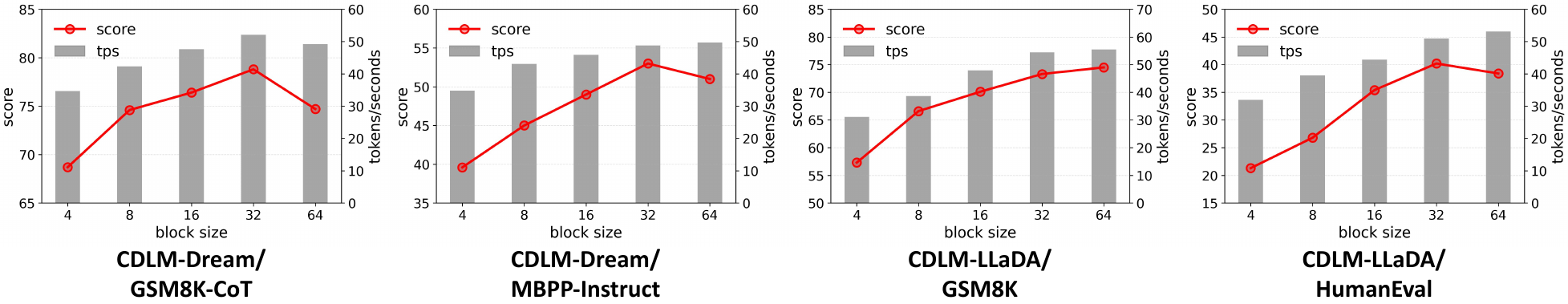}
  \vspace{-7mm}
  \caption{\textbf{Impact of inference-time block size.}
   Accuracy and throughput (TPS) across different block sizes $B\in\{4,8,16,32,64\}$ for CDLM--Dream and CDLM--LLaDA.
   All models are trained with $B{=}32$ and evaluated by varying the inference-time block size.
  }
  \label{fig:appendix-block-size}
\end{figure*}

\subsection{Validation Trends During Training}
\label{appendix:validation-trend}

We track validation performance and average refinement iterations across training epochs
for CDLM--Dream, following the training setup in Table~\ref{tab:cfg-dream}.
Figure~\ref{fig:appendix-validation} shows the trends on GSM8K-CoT (left) and MBPP-Instruct (right).
On GSM8K-CoT, validation accuracy steadily improves in early epochs and then saturates,
while the average number of refinement iterations consistently decreases.
This indicates that training encourages the model to reach correct solutions in fewer steps,
aligning with our goal of accelerating inference.
On MBPP-Instruct, validation accuracy exhibits moderate fluctuations and a downward trend
in later epochs, whereas the average number of refinement iterations continues to decrease.
This suggests that continued training further reduces the number of refinement iterations,
but may also bias the model toward math-style reasoning due to the composition of the training data.
We therefore select checkpoints before overfitting becomes pronounced.

\begin{table}[t]
\caption{\textbf{Ablation of the confidence threshold.}
Performance under different token-level confidence thresholds $\tau_{\text{conf}}$
for CDLM--Dream on GSM8K-CoT (8-shot) and HumanEval-Instruct (0-shot).}
\label{tab:ablation-threshold}
\vspace{1mm}
\centering
\small
\setlength{\tabcolsep}{6pt}
\renewcommand{\arraystretch}{1.15}
\begin{tabular}{lcccc}
\toprule
\textbf{Benchmark} & $\boldsymbol{\tau_{\text{conf}}}$ & \textbf{TPS $\uparrow$} & \textbf{Latency (s) $\downarrow$} & \textbf{Score $\uparrow$} \\
\midrule
\multirow{3}{*}{\makecell[l]{\textbf{GSM8K}\\\textbf{-CoT}\\\small(8-shot)}}
  & 0.95 & 42.7 & 2.5 & \textbf{78.8} \\
  & 0.90 & 51.7 & 2.1 & \textbf{78.8} \\
  & 0.85 & \textbf{57.7} & \textbf{1.9} & 78.4 \\
\specialrule{0.9pt}{0pt}{0pt}
\addlinespace[2pt]
\multirow{3}{*}{\makecell[l]{\textbf{HumanEval}\\\textbf{-Instruct}\\\small(0-shot)}}
  & 0.95 & 34.4 & 2.8 & \textbf{51.2} \\
  & 0.90 & 43.3 & 2.2 & 50.0 \\
  & 0.85 & \textbf{47.3} & \textbf{2.0} & 48.2 \\
\bottomrule
\end{tabular}
\vskip -0.05in
\vspace{-0.1in}
\end{table}

\subsection{Token-Level Confidence Threshold}
\label{appendix:conf-thres}

Table~\ref{tab:ablation-threshold} sweeps the token-level confidence threshold $\tau_{\text{conf}}\!\in\!\{0.85,0.90,0.95\}$ to characterize conservative-aggressive decoding behavior.
Increasing $\tau_{\text{conf}}$ makes finalization more conservative, tokens are only accepted when the model is highly confident, so throughput (TPS) decreases and per-sample latency increases.
Decreasing $\tau_{\text{conf}}$ makes decoding more aggressive and has the opposite effect.
This monotonic speed trend holds on both GSM8K-CoT and HumanEval-Instruct.
Concretely, on GSM8K-CoT, TPS/latency moves from $42.7/2.5$\,s at $\tau{=}0.95$ to $57.7/1.9$\,s at $\tau{=}0.85$; on HumanEval-Instruct, from $34.4/2.8$\,s to $47.3/2.0$\,s.

Accuracy shows a similar monotonic trend in $\tau_{\text{conf}}$.
On GSM8K-CoT, scores are essentially flat at $78.8$ for $\tau{=}0.95$ and $0.90$, with a slight drop at $\tau{=}0.85$ ($78.4$).
Similarly, on HumanEval-Instruct, the best score appears at the most conservative setting ($51.2$ at $\tau{=}0.95$), 3 points higher than the most aggressive setting ($\tau{=}0.85$).
These results suggest that raising $\tau_{\text{conf}}$ trades speed for quality, but the gain is not linear.
In practice, we find $\tau_{\text{conf}}{=}0.90$ to be a robust default that balances speed and quality across tasks.

\subsection{Sensitivity Analysis on Inference Block Size}
\label{appendix:inference-block-size}

We study the effect of varying the inference-time block size by evaluating CDLM models trained with $B{=}32$ under block sizes $B\in\{4,8,16,32,64\}$.
Figure~\ref{fig:appendix-block-size} reports accuracy and throughput (TPS) on representative benchmarks.

Across both CDLM--Dream and CDLM--LLaDA, throughput increases monotonically as the block size grows from $B{=}4$ to $B{=}32$, reflecting more aggressive exploitation of parallelism and a gradual move away from AR-style decoding.
However, increasing the block size beyond the training configuration, from $B{=}32$ to $B{=}64$, does not consistently yield further speedups.
Instead, it often results in saturated or even reduced TPS, indicating a train--inference discrepancy because the models are not trained to predict and complete such large blocks.

For most tasks, the best accuracy is achieved when the inference-time block size matches the training block size ($B{=}32$), suggesting that aligning training and inference configurations helps avoid performance degradation.
To support a wider range of block sizes at inference time, future work could explore curriculum-based training schemes that begin with smaller block sizes (e.g., $B{=}8$ or $16$) and gradually increase the difficulty of completing multiple blocks.

\begin{figure*}[t]
  \centering
  \includegraphics[width=\textwidth]{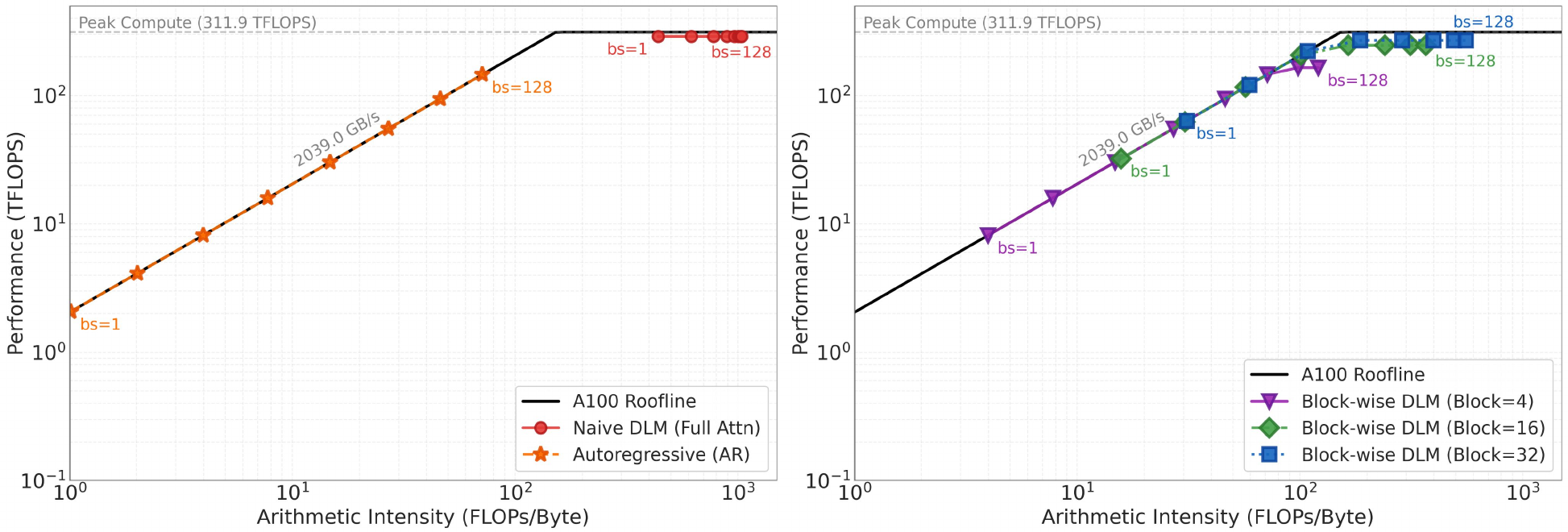}
  \vspace{-7mm}
  \caption{\textbf{Roofline analysis of batched inference across language model variants.}
  Roofline placement of decoding in autoregressive (AR) models, vanilla DLMs, and block-wise DLMs with block sizes $B\in\{4,16,32\}$ under batch sizes $\mathrm{bs}\in\{1,2,4,8,16,32,64,128\}$.
  All results are obtained via roofline simulation assuming an NVIDIA A100-SXM4-80GB GPU with peak FP16 performance of 311.9 TFLOP/s and memory bandwidth of 2039.0 GB/s.}
  \label{fig:appendix-roofline}
\end{figure*}

\subsection{Roofline Analysis of AR Models, Vanilla DLMs, and Block-wise DLMs}
\label{appendix:roofline-analysis}

Following Section~\ref{subsubsec:system-analysis}, we also perform a roofline analysis to characterize the hardware efficiency of AR decoding, full-attention vanilla DLMs, and block-wise DLMs (i.e., CDLM).
We derive roofline parameters for an NVIDIA A100-SXM4-80GB (GA100) GPU~\cite{nvidia_ampere_whitepaper, nvidia_a100_datasheet}, assuming dense FP16 Tensor Core computation (without 2:4 structured sparsity) at the boost clock frequency.
The theoretical peak FP16 performance is
\[
\begin{aligned}
\text{Peak FP16 (dense)}
&= 108~\mathrm{SM}
   \times 4~\mathrm{Tensor~Core/SM} \\
&\quad \times 256~\mathrm{FMA/(cycle\cdot Tensor~Core)} \\
&\quad \times 1.41~\mathrm{GHz} \times 2~\mathrm{FLOP/FMA} \\
&\approx 311.9~\mathrm{TFLOP/s}.
\end{aligned}
\]
With memory bandwidth of $2039.0~\mathrm{GB/s}$, the theoretical ridge point is
\[
\mathrm{AI}_{\mathrm{ridge}} = \frac{311.9~\mathrm{TFLOP/s}}{2039.0~\mathrm{GB/s}} \approx 153.0~\mathrm{FLOP/Byte}.
\]
Note that the AR and DLM backbones differ (e.g., GQA vs. MHA), so absolute AI values are not directly comparable across model families.

\paragraph{AR Models and Vanilla DLMs.}
In Figure~\ref{fig:appendix-roofline} (left), AR models start in a strongly memory-bound regime, with AI near 1 at $\mathrm{bs}=1$.
Even at $\mathrm{bs}=128$, AI remains well below the ridge point, and performance scales noticeably with batch size.
In contrast, vanilla DLMs are compute-bound even at $\mathrm{bs}=1$, with AI ($\approx 438.9$) far exceeding the ridge point.
Accordingly, performance quickly saturates and shows little improvement with increasing batch size.
The observed plateau lies slightly below the theoretical peak, as non-linear operations such as layer normalization and softmax are executed on vector units with lower peak throughput than Tensor Cores.
This heterogeneity in execution units results in a lower effective performance ceiling during inference.

\paragraph{Block-wise DLMs.}
Figure~\ref{fig:appendix-roofline} (right) shows the block-wise DLMs with block sizes $B=4,16,32$.
At $\mathrm{bs}=1$, their AI values (4.0, 15.8, and 31.1, respectively) are higher than AR decoding but lower than vanilla DLMs.
This reflects increased intra-block data reuse: processing multiple tokens within a shared KV-cache context and reusing the same model weights across tokens in the block amortizes memory access and raises AI.
For small batch sizes, block-wise DLMs remain memory-bound and exhibit strong scaling as batch size increases.
As $B$ grows, performance tends to saturate around $\mathrm{bs}=64$ for $B=4$, $\mathrm{bs}=16$ for $B=16$, and $\mathrm{bs}=8$ for $B=32$.

Overall, block-wise DLMs occupy a middle ground between AR models, which are memory-bound, and vanilla DLMs, which are compute-bound.
By exploiting parallelism within blocks while retaining moderate AI, block-wise decoding improves compute utilization in low-batch regimes without immediately saturating available compute.

\section{Limitations and Future Directions}
\label{appendix:future}

Our current training relies on offline, static trajectories. 
Although we carefully select datasets to limit domain bias, the student can still overfit to the teacher’s priors because supervision does not adapt during training.
A straightforward direction is to broaden the data mixture and scale beyond our current $\sim$15k samples to cover more diverse domains.

A complementary path is to move from offline supervision to on-the-fly generation.
If the student generates trajectories during training and the teacher verifies or refines them online, on-policy learning could reduce the training--inference discrepancy.
In practice, this is currently constrained by DLM sampling speed: slow generation limits the feasibility of closed-loop training without prior trajectory materialization. 
Improving DLM inference throughput is therefore a prerequisite for large-scale online training.

Another promising direction is to use CDLM as a draft model within speculative decoding frameworks~\cite{leviathan2023fast}, where diffusion models can serve as fast proposers for autoregressive verifiers.
A naive DLM drafter yields limited benefit because standard DLMs require many refinement steps to produce coherent text. In contrast, a \emph{consistency}-trained DLM (CDLM) can generate reasonable drafts in far fewer steps.
Pairing a CDLM drafter with an autoregressive (AR) verifier (cf.\ \cite{christopher2025speculative}) is promising: the CDLM proposes multi-token candidates quickly, and the AR model supplies strong final validation.

CDLM’s performance is ultimately bounded by the teacher: a bidirectional DLM distilled into a block-causal student cannot exceed the teacher’s knowledge. 
To lift this ceiling, distillation from stronger autoregressive (AR) teachers is a natural next step.
In this work we intentionally used high-quality, third-party-curated Qwen2.5-7B generations~\cite{lansechen_easy_2025, lansechen_hard_2025} as ground-truth text, hoping to inject AR knowledge into the student. 
A more direct, and likely more effective, approach is to distill from an AR teacher: AR models are already scaled up, typically stronger, and much faster at generation than naive DLMs, making large-scale dataset construction easier.
Identifying effective ways to extract suitable training trajectories from AR models is therefore a key prerequisite.


\end{document}